\documentclass[letterpaper]{article} 
\usepackage[submission]{aaai24}  
\usepackage{times}  
\usepackage{helvet}  
\usepackage{courier}  
\usepackage[hyphens]{url}  
\usepackage{comment}
\usepackage{graphicx} 
\usepackage{natbib}  
\usepackage{caption} 
\usepackage{algorithm}

\usepackage{algpseudocode}
\usepackage{multirow}

\usepackage{xcolor}
\usepackage{amsmath, amsthm, amssymb}

\usepackage{subfig}
\usepackage{subfloat}
\newtheorem{prop}{Proposition}

\usepackage{newfloat}
\usepackage{listings}
\DeclareCaptionStyle{ruled}{labelfont=normalfont,labelsep=colon,strut=off} 
\floatstyle{ruled}
\newfloat{listing}{tb}{lst}{}
\floatname{listing}{Listing}
\title{My Publication Title --- Multiple Authors}
\author {
    Xuan Zhao\textsuperscript{\rm 1},
    Simone Fabbrizzi\textsuperscript{\rm 2},
    Paula Reyero Lobo\textsuperscript{\rm 3},
    Siamak Ghodsi\textsuperscript{\rm 4},
    Klaus Broelemann\textsuperscript{\rm 1},
    Steffen	Staab \textsuperscript{\rm 5},
    Gjergji	Kasneci \textsuperscript{\rm 6}    
}
\affiliations {
    \textsuperscript{\rm 1}SCHUFA Holding AG, Germany\\
    \textsuperscript{\rm 2}Leibniz Universität Hannover, Germany\\
    \textsuperscript{\rm 3}The Open University, United Kingdom\\
    \textsuperscript{\rm 4}L3S research center Hannover, Germany\\
    \textsuperscript{\rm 5}University of Stuttgart\\
    \textsuperscript{\rm 6}Technical University of Munich\\

}
\title{Adversarial Reweighting Guided by Wasserstein Distance for Bias Mitigation}

\usepackage{bibentry}

\begin{document}

\maketitle

\begin{abstract}
The unequal representation of different groups in a sample population can lead to discrimination of minority groups when machine learning models make automated decisions.
To address these issues, fairness-aware machine learning jointly optimizes two (or more) metrics aiming at predictive effectiveness and low unfairness. However, the inherent under-representation of minorities in the data makes the disparate treatment of subpopulations less noticeable and difficult to deal with during learning. 
In this paper, we propose a novel adversarial reweighting method to address such \emph{representation bias}. 
To balance the data distribution between the majority and the minority groups, our approach deemphasizes samples from the majority group.
To minimize empirical risk, our method prefers
 samples from the majority group that are close to the minority group as evaluated by the Wasserstein distance. Our theoretical analysis shows  the effectiveness of our adversarial reweighting approach. Experiments demonstrate  that our approach mitigates bias without sacrificing classification accuracy, outperforming related state-of-the-art methods on image and tabular benchmark datasets.
\end{abstract}

\section{Introduction}\label{sec:intro}
Machine learning models that are trained on personal data may discriminate against groups with sensitive attributes. Broadly speaking, there are three major paradigms to address this problem. The first paradigm assumes 
 fairness can be measured. Then, the minimization of unfairness metrics is integrated in the empirical risk minimization as a multi-objective optimization problem \cite{aghaei2019,berk2017}.  The second paradigm assumes that  discrimination arises from the use of protected (i.e., sensitive) attributes and those correlated to them. Therefore, removing sensitive information from the input data can support learning fair models \cite{creager2019}. The third paradigm builds on the assumption that discrimination arises from biased labeling processes (e.g., through biased domain knowledge or biased human feedback). Corresponding approaches aim at identifying and correcting label bias~\cite{jiang2019}, such as the adaptive sensitive reweighting of instances~\cite{krasanakis2018}.

These paradigms do not deal directly with the issue 
that, by definition, minority groups are smaller than the majority. The effects of 
 under-represented data samples in the learning process are `overridden' by the prevalence of data samples from the majority group. The under-representation negatively affects the sensitivity of the fairness metrics and can hide undesirable correlations between attributes in the minority group. That is, leading to a \emph{representation bias}. 
 
 We propose a reweighting scheme to mitigate predictive quality issues arising from the imbalance between sensitive groups. We do so by mapping the data into a latent space where the data distribution becomes non-discriminatory with respect to the sensitive attribute. Simultaneously, the empirical risk for the classification task at hand is minimized. Our method addresses representation bias by weighting the samples from the majority group. It aims to maintain the class-wise discriminatory information of the data samples from the majority group that are further away from the minority group, but downplay their importance. Hence, the majority and minority groups become similar in distribution and almost non-discriminatory in classification. We use the critic of a Wasserstein Generative Adversarial Network (WGAN) with gradient penalty \cite{gulrajani2017} to approximate distances between samples from the minority and reweighted majority groups in the latent space. 


The rationale for our method is that if subgroups are sufficiently represented in a non-discriminatory way, \textit{bias in prediction} would be substantially reduced, if not eliminated \cite{chai2022}. Reweighting instances has been adopted in methods for learning from imbalanced datasets \cite{bao2020,zhang2019}, 
which focus on optimizing the performance under a class imbalance, without considering representation bias. Our method is different from existing adversarial methods \cite{adel2019a,madras2018,wadsworth2018} in exploiting the competition between the reweighing component and the discriminator of the GAN framework, as an additional discriminator has been generally used to decorrelate feature embeddings from sensitive information. 

We perform experiments on different datasets and compare with four 
state-of-the-art fairness-aware methods. 
Our method outperforms its competitors in mitigating bias while maintaining high prediction quality, as demonstrated by the experimental evaluation of image and tabular benchmark datasets.
Hence, our method inherently addresses fairness as well as prediction quality issues that might arise from learning on imbalanced datasets with respect to sensitive groups.

We summarize our contribution as follows:
(1) We formulate a novel data transformation and sample-based reweighting method for mitigating representation bias related to sensitive groups in classification tasks.
(2) We show theoretically that by closing the Wasserstein distance gap between sensitive groups in the latent space during training, our reweighting approach leads to predictions that adhere to demographic parity. 
(3) We provide a thorough evaluation of the proposed technique on image and tabular benchmark datasets and show the viability of our approach with respect to robustness to fairness, accuracy and label noise. Code is available at \url{https://anonymous.4open.science/r/wasserstein_reweight-46E6/}.


\section{Related Work} \label{sec:related}

\subsubsection{Adversarial methods}
Existing adversarial fairness methods \cite{adel2019a,madras2018,wadsworth2018} use in an in-processing fashion a discriminator
to decorrelate the embeddings and the sensitive attribute. The authors of \cite{choi2019} and \cite{kim2019} propose to minimize mutual information between the biased labels and the embedding through adversarial training. These works disentangle the sensitive attribute in the latent space, yet they do not consider the under-representation of sensitive groups. Our work considers representation bias in the decorrelation process by reweighting to align the distributions of the sensitive groups instead of only adjusting the encoder. 



\subsubsection{Reweighting methods}

Fairness with Adaptive Weights \cite{chai2022} also constrains the sum of weights among sensitive groups to be equal, assigning weights to a sample based on its misclassification likelihood. Adaptive sensitive reweighting to mitigate bias \cite{krasanakis2018} assigns weights to samples based on their alignment with the unobserved true labeling. 
As highlighted example, Adversarial reweighting for domain adaptation \cite{gu2021} aims to align the distributions of the source and target domains, yet it deals with the domain adaptation problem. In our work, we extend the concept of reweighting based on the Wasserstein distance to the fairness domain.

\subsubsection{Imbalanced classification}
There are two main imbalanced classification methods: resampling and cost-sensitive learning. Resampling methods achieve balance between class groups by oversampling the group with a small size (the minority group in fairness settings) or undersampling the group with a large size (the majority group in fairness settings) or both.
For instance, \cite{bao2020} carries out classification using clustering centers in latent space to balance among the groups, which is equivalent to undersampling all groups.
Cost-sensitive learning assigns higher weights to samples from groups with small sizes during training such that the costs of misclassifying these samples are higher than that from groups with large size. There are various methods on such weighting schemes. For example, in \cite{huang2019}, the authors balance the representations of groups by constraining the embedding to keep inter-cluster margins both within and between classes. Note that these works deal with class imbalance while our work focuses on imbalance regarding sensitive attributes.

\section{Background}\label{sec:back}

Throughout this work, we consider binary classifiers that produce
 estimations $\hat{y}\in\{0,1\}$ 
 \newcommand{\dataset}{\mathcal{D}}
 for a given a dataset $\dataset=\{(x_1,y_1),\ldots,(x_n,y_n)|x_i\in X \subseteq\mathbb{R}^d, y_i\in Y=\{0,1\}\}$, where $x_i$ represents vectors of attributes, and $y_i$ is the target label of data instance $i$.
 Let the first component of $x_i$ describe the sensitive attribute $s_i=x_{i,1}\in \{0,1\}$.
 The values of the sensitive attributes $s_i$ distinguish between the majority group  having $n_p$ many samples and the minority (i.e., sensitive or under-represented) group having $n_u$ many samples. 
Without loss of generality, we assume that $\forall i: 1\leqslant i \leqslant n_p  \Rightarrow s_i=1.
\forall i: n_p+1\leqslant i \leqslant n  \Rightarrow s_i=0$.
 

 \subsection{Fairness notions} \label{sec:notion}

Disparate treatment \cite{zafar2017} occurs when the classifier makes different predictions on individuals from different sensitive groups when the input features are identical. To mitigate it, the classifier should achieve calibration across the sensitive groups: $P(\hat{y}|x, s) = P(\hat{y}|x)$.
Disparate impact \cite{kamiran2012}  evaluates the difference in positive outcome rate between groups and is eliminated when the predictive outcome $\hat{y}$ is independent of $s$: $P(\hat{y}|s = 0) = P(\hat{y}|s = 1)$. Nevertheless, eliminating disparate impact does not ensure a fair classifier. Since the sample distribution among sensitive groups is not naturally even, the classifier might focus on the majority group while ignoring decisions on the minority group. Furthermore, even if zero disparate impact is achieved, we might sacrifice the classifier's performance since statistical features of different sensitive groups usually vary. Disparate mistreatment \cite{hardt2016} occurs when the misclassification rates (false positives and false negatives) of different sensitive groups are different. In this case, the measurement of disparate mistreatment requires labeled data.
Earlier works, including \cite{chouldechova2016}, state that there is usually tension among the disparate mistreatment criteria. 
Disparate FPR (false positive rate) and Disparate FNR (false negative rate) are commonly used to reduce disparate mistreatment: $P(\hat{y}\neq y|y=1, s) = P(\hat{y}\neq y|y=1)$ and  $P(\hat{y}\neq y|y=0, s) = P(\hat{y}\neq y|y=0)$.

\subsection{Wasserstein distance methods}

\textbf{Definition} The Wasserstein distance between two distributions $\mu$ and $\nu$ is defined by $W(\mu, \nu) =\text{min}_{\pi \in \Pi}\mathbb{E}_{(x,x')\sim\pi}[\|x-x'\|^p]$, where $\Pi$ is the set of couplings of $\mu$ and $\nu$, i.e., $\Pi = \{\pi|\int \pi(x,x')dx'=\mu(x)$,$\int \pi(x,x')dx=\nu(x')\}$, and $p\geqslant 1$. In the later sections of this work, $p = 2$. 
Following the Kantorovich-Rubinstein duality, we have the dual form of the Wasserstein distance of $W(\mu, \nu) = \text{max}_{\|f\|_{L}\leqslant1}\mathbb{E}_{x\sim\mu}[f(x)]-\mathbb{E}_{x'\sim\nu}[f(x')]$, where the maximization is over all 1-Lipschitz functions $f$: $\mathbb{R}^d \rightarrow\mathbb{R}$. 

\subsubsection{Fairness-aware classification}

Wasserstein distance, also known as Optimal Transport (OT) distance, is a metric in the space of measures with finite moments that can be used to evaluate how two distributions are different from one another. A valuable application of the properties of this metric is the computation of the barycenter of two distributions. Such technique has been leveraged in fairness-aware classification methods \cite{Zehlike20,jiang2019a} to enforce statistical parity. Notably, Wasserstein Fair Classification (WFC) \cite{jiang2019a} quantile matches the predictions of the sensitive group to the predictions of the barycenter of all groups.  
Fairness with Continuous Optimal Transport \cite{chiappa2021} introduces a stochastic-gradient fairness method based on a dual formulation of continuous OT instead of discrete OT to improve performance.

\subsubsection{Generative methods}\label{sec:generative} Wasserstein GANs (WGANs) \cite{arjovsky2017} are based on minimising the Wasserstein distance between a real and a generated distribution by weight clipping to enforce a Lipschitz constraint on the critic, improving the performance of plain GANs. WGANs with Gradient Penalty (WGAN-GP) \cite{gulrajani2017} is a relaxed version of the Lipschitz constraint, which follows that functions are 1-Lipschitz if the gradients are of norm at most 1 everywhere.  

\begin{figure}
    \centering
    \includegraphics[width=.49\textwidth]{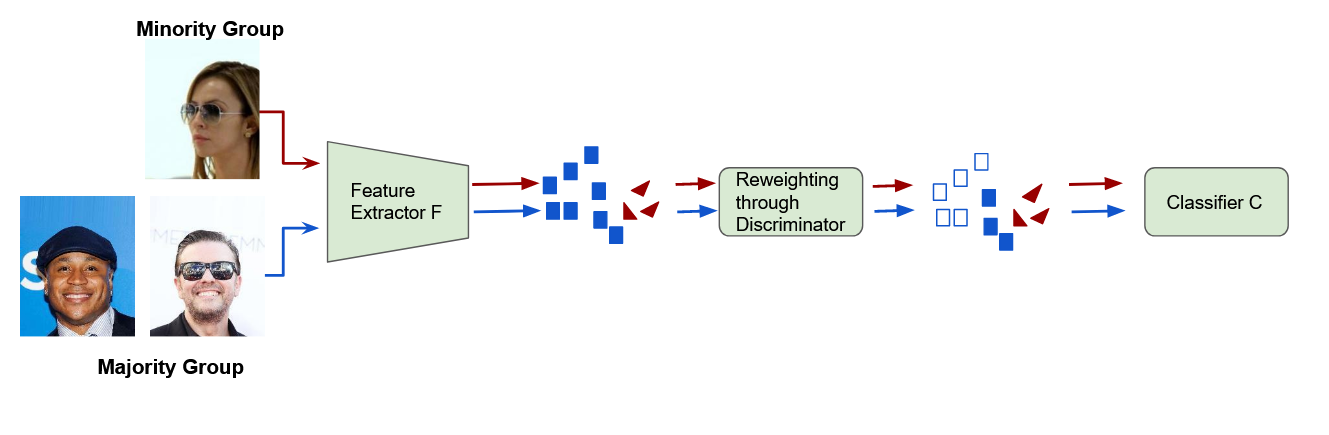}
    \caption{Architecture of our approach. The arrows show the computational flow for the minority (resp. majority) group in the classification task (e.g., predicting whether a person in the image is wearing a hat). Representation bias is indicated by blue and red triangles. Both minority and majority groups are mapped onto a latent space by the feature extractor. Then, majority group instances are reweighted to match the minority group distribution, aiming to decrease the distance with respect to the sensitive attribute.
}
    \label{fig:pipeline}
\end{figure}

\section{Our Adversarial Reweighting Approach}\label{sec:method}

\subsection{Problem formulation}\label{sec:original}

Consider a feature extractor $F_\phi : X \rightarrow Z\subseteq \mathbb{R}^k$ that maps raw data from the dataset $\dataset$ into a latent feature space. 
The transformation function can be viewed as an embedding component for more effective comparison of instances in latent space. A binary classifier $C_\theta : Z \rightarrow Y$ with parameters $\theta$  maps the results of the transformation $F_\phi(x)$ to a binary label $\hat{y} \in Y = \{0, 1\}$. For simplicity, we demonstrate our method in the scenario with binary sensitive attribute. However, it is straight-forward to extend our method to handle a multi-categorical sensitive attribute or multi-sensitive attributes (see Appendices). 

As part of our problem definition, we indicate the training objective that minimizes the weighted empirical risk  as:
\begin{equation} \label{eq:vanilla}
\underset{\theta}{\text{min}}\sum_{i=1}^{n}w_i\mathcal{L}(y_i,(C_{\theta}\circ F_\phi)(x_i)),  \text{ with } w_i\geqslant 0
\end{equation}

 with $\mathcal{L}$ representing cross-entropy loss.
The overall pipeline we seek to develop is illustrated in Figure~\ref{fig:pipeline}. The feature extractor is optional and may not be needed given low dimensionality of the data.

If all training samples receive the same weights, the classifier will tend to focus more on the majority group leading to representation bias. 
We seek to maintain the weights for samples from the minority group (i.e., $\forall i: s_i=0 \Rightarrow w_i=1$) while lowering the weights for samples from the majority group, such that the sum of weights is the same for both groups: 
\begin{equation}\label{eq:sameweight}
    \sum_{i=1}^{n_p}w_i=n_u
\end{equation}
To avoid information loss by assigning zero weights to some samples from the majority group, we introduce a regularization constraint  to our risk minimization term:
\begin{equation}\label{eq:balance}
    \sum_{i=1}^{n_p}(w_i-\frac{n_u}{n_p})^2\leqslant Tn_u
\end{equation}

The sum is minimal (namely zero) if $\forall i: w_i=\frac{n_u}{n_p}$. Thus,
by adjusting the value of $T$ 
we can balance between similarity and dissimilarity of the weights of samples from the majority group.

Together, Equations~\eqref{eq:vanilla}, \eqref{eq:sameweight}, and~\eqref{eq:balance} constrain the problem space. On their own, however, they do not fully account for within and between sample group differences, thus not always improving group-based fairness metrics. The problem is now defined to seek a weighting scheme that fulfills
the Equations~\eqref{eq:vanilla}, \eqref{eq:sameweight}, and~\eqref{eq:balance} while mitigating the representation bias in a robust way. 

\subsection{Adversarial reweighting} 
The goal of our weighting scheme is to determine weights such that the majority and minority group weighted distributions become similar and, hence, the classifier is less prone to biased and unfair predictions.
Our adversarial learning of data weights in the majority group targets to pay more attention to samples in the majority group that are closer to the minority group during the training, without completely loosing the information contained in other samples of the majority group.
We measure the similarity of weighted distributions by the Wasserstein distance in the latent space for the reason that 
 approximating the Wasserstein distance in the latent space is computationally less demanding in a low-dimensional space. 
 
 In the following, we first show that enforcing a small Wasserstein distance in the latent space ensures small distance in the prediction space. Then, we discuss the detailed adversarial reweighting model.
 

\subsubsection{Theoretical proof of enforcing demographic parity}\label{sec:proof} 


We show that enforcing the Wasserstein distance, indicated as $W(\cdot, \cdot)$, being small in the latent space enforces it to be small in the prediction space as well.

\begin{prop}
\label{prop:Lipschitz condition}

Given two measures $\mu$ and $\nu$ over a metric space $(Z, d_Z)$ and a $K$-Lipschitz function $C:(Z, d_Z) \rightarrow (Y, d_Y)$, we have that $$W(C_{\#}\mu, C_{\#}\nu) \leq K\cdot W(\mu, \nu)$$
Where $C_{\#}\mu$ is the push-forward measure along the function $C$. For details of proof, please refer to Appendices. 
\end{prop}

Note that, since we deal with classifiers over a finite dataset, the $K$-Lipschitz condition for a (binary) classifier C: $Z \rightarrow \{ 0, 1\}$ amounts to asking that for every $z$ and $z'$ such that $C(z) \neq C(z')$ we have that $\frac{1}{K} \leq  d_Z(z,z')$ because the set $\{0, 1\}$ is endowed with the discrete metric
. Given that we consider only finite datasets, we can always find such a $K$. Note that this result is only valid for a given dataset and it does not generalize unless we assume that the condition $\frac{1}{K} \leq  d_Z(z,z')$ for every $z$ and $z'$ such that $C(z) \neq C(z')$  holds true for the new data as well.

Thus, we have that if the Wasserstein distance is close to 0 in the latent space, it will be close to 0 in the prediction space. Note that $W(C_{\#}\mu, C_{\#}\nu) = 0$ means that $C_{\#}\mu = C_{\#}\nu$ and this implies demographic parity. Indeed, defining the distribution $\zeta := \frac{1}{2}\mu + \frac{1}{2}\nu$ describes the probability of being sampled from either the majority or the minority groups. Then, since $\mu$ and $\nu$ are discrete, we have that $C_{\#}\zeta = \frac{1}{2}C_{\#}\mu + \frac{1}{2}C_{\#}\nu$. Therefore,  $C_{\#}\mu = C_{\#}\nu$ implies that the probability of $C(z) = 1$ is irrespective of the fact that $z$ is sampled from the majority or minority groups.

\subsubsection{Adversarial reweighting model}\label{sec:adver}
%
%
 
We approximate the computation of the Wasserstein distance by a neural network discriminator $D$ using the  gradient penalty
technique of WGAN-GP \cite{gulrajani2017}:

\begin{equation} \label{eq:wass}
W(\mu,\nu)\approx\underset{\theta_D}{\text{max}}\left(\mathbb{E}_{z\sim\mu}[D(z;\theta_D)]-\mathbb{E}_{z'\sim\nu}[D(z';\theta_D)]\right)
\end{equation}

%
We define the (weighted) empirical distributions of the minority group $\mathcal{P}_U$ and the majority group $\mathcal{P}_P(\textbf{w})$ using the Dirac delta function $\delta(\cdot)$ as:
\begin{eqnarray}
\mathcal{P}_U =\frac{1}{n_u}\sum_{i=n_p+1}^{n}\delta(F(x_i)),\\
\mathcal{P}_P(\textbf{w}) =\frac{1}{n_u}\sum_{i=1}^{n_p}w_i\delta(F(x_i)),\text{ with }\sum_{i=1}^{n_p}w_i=n_u
\end{eqnarray}


Then, we optimize the weights by minimizing the Wasserstein distance between the minority and reweighted majority distributions, whereby  Equations~\eqref{eq:sameweight} and~\eqref{eq:balance} define the solution space for the weights $\mathcal{W} = \{\textbf{w} :\textbf{w}= (w_1,w_2,...,w_{n_p})^T, w_i\geqslant0, \sum_{i=1}^{n_p}w_i=n_u, \sum_{i=1}^{n_p}(w_i-\frac{n_u}{n_p})^2\leqslant Tn_u\}$: 
\begin{equation} \label{eq:minw}
\underset{\textbf{w}\in\mathcal{W}}{\text{min}}W(\mathcal{P}_U,\mathcal{P}_P(\textbf{w}))
\end{equation}

Because of Proposition \ref{prop:Lipschitz condition}
, we know that such minimization contributes to reducing the disparity between majority and minority groups.

%
%
%

If $f$ is a measurable function and $\mu=\sum\alpha_i \delta(x_i)$ a discrete distribution, we have that $f_\#\mu = \sum\alpha_i \delta(f(x_i))$. Hence, combining Equations ~\eqref{eq:wass} and ~\eqref{eq:minw} results into a minmax problem, yields:

\begin{equation} \label{eq:final}
\underset{\mathbf{w}\in\mathcal{W}}{\text{min}}\underset{\theta_D}{\text{max}}\bigg(\sum_{i=1}^{n_p}w_iD(z_i^p;\theta_D)
-\sum_{i=1}^{n_u} D(z_i^u;\theta_D)\bigg)
\end{equation}
In Equation \eqref{eq:final}, the discriminator is trained to maximize the average of its outputs on the minority and majority group; adversarially, the weights for samples from the majority group are learned to minimize the (reweighted) average of the outputs of the discriminator. As a result, the samples from the majority group with smaller discriminator
outputs (closer to the minority group) will be assigned higher weights. Therefore, defining the reweighted cross-entropy loss on the (reweighted) data distribution in Equation \eqref{eq:vanilla} mitigates the representation bias regarding the minority groups.

\subsection{Training algorithm}
To train the feature extractor $F_\phi$ and the classifier network $C_\theta$, 
the network parameters ($\phi$, $\theta$) and learn the weights $w$ with $D$ are updated by fixing others. We alternately train the following two steps.

\textbf{Updating $\phi$ and $\theta$ while fixing $\mathbf{w}$ and $\theta_D$.} 
$\phi$ and $\theta$ are updated to minimize the loss in Equation \eqref{eq:vanilla} for $S$ steps batch-wise while $\mathbf{w}$ and $\theta_D$ are fixed.

\textbf{Updating $\mathbf{w}$ and $\theta_D$ while fixing $\phi$ and $\theta$.}
Embeddings of training data on both majority and minority groups are acquired through the feature extractor $F$ while $\phi$ and $\theta$ are fixed. $\mathbf{w}$ in Equation \eqref{eq:final} is learned: Equation \eqref{eq:final} is a min-max optimization problem, the weights $\mathbf{w}$ and the parameters $\theta_D$ of the discriminator could be optimized alternatively. 
We could first fix $w_i=\frac{n_u}{n_p}$ for all $i$ and optimize $\theta_D$ to maximize the objective function in Equation \eqref{eq:final} using the gradient penalty technique, as in WGAN-GP \cite{gulrajani2017}. Then, fixing the discriminator, we optimize $\mathbf{w}$. We denote $d_i=D(F_\theta(x_i);\theta_D)$ and $\mathbf{d}=(d_1, d_2, ...d_{n_p})^T$. The optimization problem for $\mathbf{w}$ becomes a constrained least squares problem:

\begin{equation} \label{eq:w}
\underset{\mathbf{w}}{\text{min}}\,\mathbf{d}^T\mathbf{w},  s.t.  w_i\geqslant0, \sum_{i=1}^{n_p}w_i=n_u,\sum_{i=1}^{n_p}(w_i-\frac{n_u}{n_p})^2\leqslant Tn_u
\end{equation}

\section{Experiments}\label{sec:eva}

We evaluate the performance of our reweighting approach on three benchmark datasets comparing it against eight methods using four different metrics: Accuracy, Disparate Impact, Disparate FPR and Disparate FPR.

Four fairness-agnostic methods help us to better understand issues with unfairness. Inspired by \cite{chai2022} we compare against  (1) \textbf{Baseline} (Neural Network (NN) based classification without fairness constraints); 
(2) Simple \textbf{Reweighing}: NN classification with assigning same balancing weights to samples the majority group;
(3) \textbf{Undersampling} forms the training dataset by
balancing group sizes via undersampling from the majority group; (4) \textbf{Oversampling} balances group sizes by repeating sampling from the minority group.

We choose four further competing methods mentioned in earlier sections: (5) Adaptive sensitive reweighting (\textbf{ASR}) reweights samples to balance target class occurrences. (6) Wasserstein fair classification (\textbf{WFC}) matches quantiles of the predictive distribution of the sensitive group to the all-group Wasserstein barycenter. (7) 
The Fair Adversarial Discriminative (\textbf{FAD}) model \cite{adel2019a} decorrelates the sensitive information from the embeddings by adjusting the encoding/feature extraction process using adversarial training. (8) Fairness with Adaptive Weights (\textbf{FAW}) constrains the sum of weights. They (i) are designed to address bias, (ii)  follow conceptually similar strategies, and (iii) can also be flexibly applied to different modalities (tabular and images).

Our networks are trained on an Intel(r) Core(TM) i7-8700 CPU. The networks in our experiments are built based on Pytorch \cite{NEURIPS2019_9015} and the optimization in Equation \eqref{eq:w} is performed with the python package CVXPY \cite{diamond2016cvxpy}. 



\subsection{Data and training details}
\subsubsection{Image dataset}

We test three datasets based on CelebA \cite{liu2015faceattributes} which contain 70\% male images vs. 30\% female images, 80\% male images vs. 20\% female images and 90\% male images vs. 10\% female images, respectively. We use three different distributions for the sensitive attribute to analyze in which imbalance situation our method is more suitable. We maintain the class imbalance in the three datasets constant, namely 70\% not wearing a hat and 30\% wearing a hat. For more details on CelebA, please refer to Appendices. The classification task is to identify whether the person in the picture is wearing a hat. 

For the feature extractor $F$, we apply ResNet-18 \cite{he2016residual} architecture, pre-trained on ImageNet \cite{deng2009imagenet}, without the last fully-connected layer for simplicity. For the feature extractor $F_\phi$ and classifier $C_\theta$, we use the stocastic gradient descent (SGD) algorithm \cite{shamir2012} with a momentum of 0.9 to update $\phi$ and $\theta$. For the discriminator $D$, we use a similar architecture as the one in~\cite{gulrajani2017} with three
fully connected layers of 512, 256 and 128, 64, and 1 node, respectively; and without the last sigmoid function. We apply the Adam algorithm \cite{kingma2014method} to update $\theta_D$ with a learning rate of 0.0001. 
Following \cite{gulrajani2017}, we adjust the learning rate $\eta$ by $\eta = \frac{0.01}{(1+10p)^{-0.75}}$, where $p$ is the training progress linearly changing from 0 to 1. We set the batch size to $n_p$ 700, 800, 900, and $n_u$ 300, 200, and 100, respectively. We update $\phi$ and $\theta$
for 4 steps and then update $\theta_D$ for 1 step. Note that we choose a relatively high batch-size because estimating the true Wasserstein distance between distributions via batches requires relevant sample sizes. Details on split of training and testing datasets are shown in Appendices. 

\begin{table*}[!htbp]
  \small
  \centering
  \caption{Experiment Results on CelebA}
  \label{table:1}
  \subfloat[Experimental results of classifier (Wearing Hat) on dataset (30\% female and 70\% male)]{%
    \resizebox{\linewidth}{!}{\begin{tabular}{|c|cccc|cccc|c|}
\hline
\multirow{2}{*}{methods} & \multicolumn{4}{c|}{simple methods}                                                                                           & \multicolumn{4}{c|}{state-of-the-art methods}                                                                                              & ours                 \\ \cline{2-10} 
                         & \multicolumn{1}{c|}{baseline}    & \multicolumn{1}{c|}{reweighing}  & \multicolumn{1}{c|}{undersampling}       & oversampling & \multicolumn{1}{c|}{ASR}         & \multicolumn{1}{c|}{WFC}                & \multicolumn{1}{c|}{FAD}                 & FAW                & AR                   \\ \hline
Accuracy rate (\%)       & \multicolumn{1}{c|}{95.1 (0.7)}  & \multicolumn{1}{c|}{94.9 (0.6)}  & \multicolumn{1}{c|}{\textbf{95.3 (0.4)}} & 94.7 (0.9)   & \multicolumn{1}{c|}{93.5 (0.6)}  & \multicolumn{1}{c|}{93.6 (0.5)}         & \multicolumn{1}{c|}{\textbf{95.0 (0.7)}} & 93.7 (0.6)         & 94.7 (0.7)           \\ \hline
Disparate Impact (\%)    & \multicolumn{1}{c|}{6.0 (0.8)}   & \multicolumn{1}{c|}{6.0 (0.4)}   & \multicolumn{1}{c|}{5.7 (0.2)}           & 4.9 (0.7)    & \multicolumn{1}{c|}{5.1 (0.7)}   & \multicolumn{1}{c|}{5.3 (2.4)}          & \multicolumn{1}{c|}{5.5 (2.4)}           & \textbf{4.7 (0.4)} & \textbf{0.8 (0.5)}   \\ \hline
Disparate FPR (\%)       & \multicolumn{1}{c|}{-29.1 (1.4)} & \multicolumn{1}{c|}{-31.3 (9.2)} & \multicolumn{1}{c|}{-31.8 (8.1)}         & -24.7 (7.5)  & \multicolumn{1}{c|}{-26.2 (8.1)} & \multicolumn{1}{c|}{\textbf{4.5 (1.1)}} & \multicolumn{1}{c|}{-25.7 (5.8)}         & -13.7 (2.1)        & \textbf{-17.0 (5.1)} \\ \hline
Disparate FNR (\%)       & \multicolumn{1}{c|}{7.3 (2.9)}   & \multicolumn{1}{c|}{8.1 (4.2)}   & \multicolumn{1}{c|}{7.1 (3.6)}           & 7.9 (1.9)    & \multicolumn{1}{c|}{8.2 (1.8)}   & \multicolumn{1}{c|}{7.6 (3.9)}          & \multicolumn{1}{c|}{\textbf{6.9 (1.2)}}  & 10.0 (1.1)         & \textbf{6.5 (1.9)}   \\ \hline
\end{tabular}}%
    \hspace{.5cm}%
  }\hspace{1cm}
  \label{table:2}\subfloat[Experimental results of classifier (Wearing Hat) on the dataset (20\% female and 80\% male)]{%
    \resizebox{\linewidth}{!}{\begin{tabular}{|c|cccc|cccc|c|}
\hline
\multirow{2}{*}{methods} & \multicolumn{4}{c|}{simple methods}                                                                                          & \multicolumn{4}{c|}{state-of-the-art methods}                                                                                                & ours                \\ \cline{2-10} 
                         & \multicolumn{1}{c|}{baseline}    & \multicolumn{1}{c|}{reweighing}  & \multicolumn{1}{c|}{undersampling}      & oversampling & \multicolumn{1}{c|}{ASR}                 & \multicolumn{1}{c|}{WFC}                 & \multicolumn{1}{c|}{FAD}                 & FAW         & AR                  \\ \hline
Accuracy rate (\%)       & \multicolumn{1}{c|}{95.3 (0.9)}  & \multicolumn{1}{c|}{93.7 (0.5)}  & \multicolumn{1}{c|}{95.0 (0.4)}         & 94.9 (1.6)   & \multicolumn{1}{c|}{93.1 (0.7)}          & \multicolumn{1}{c|}{93.3 (0.8)}          & \multicolumn{1}{c|}{\textbf{95.0 (0.5)}} & 93.6 (0.8)  & \textbf{95.3 (0.9)} \\ \hline
Disparate Impact (\%)    & \multicolumn{1}{c|}{3.7 (0.7)}   & \multicolumn{1}{c|}{3.9 (0.9)}   & \multicolumn{1}{c|}{\textbf{3.1 (0.3)}} & 3.4 (0.3)    & \multicolumn{1}{c|}{4.3 (0.9)}           & \multicolumn{1}{c|}{5.1 (1.4)}           & \multicolumn{1}{c|}{18.2 (1.4)}          & 3.8 (0.4)   & \textbf{0.7 (0.4)}  \\ \hline
Disparate FPR (\%)       & \multicolumn{1}{c|}{-16.0 (1.7)} & \multicolumn{1}{c|}{-32.1 (7.3)} & \multicolumn{1}{c|}{-28.9 (4.1)}        & -17.4 (5.0)  & \multicolumn{1}{c|}{\textbf{-2.0 (4.4)}} & \multicolumn{1}{c|}{\textbf{4.0 (1.7)}}  & \multicolumn{1}{c|}{-21.0 (2.6)}         & -19.7 (1.1) & -22.2 (5.3)         \\ \hline
Disparate FNR (\%)       & \multicolumn{1}{c|}{10.3 (2.1)}  & \multicolumn{1}{c|}{16.6 (5.4)}  & \multicolumn{1}{c|}{11.9 (3.7)}         & 10.9 (3.7)   & \multicolumn{1}{c|}{\textbf{8.1 (0.8)}}  & \multicolumn{1}{c|}{\textbf{-7.2 (2.1)}} & \multicolumn{1}{c|}{10.5 (1.4)}          & 10.0 (1.1)  & 9.0 (3.7)           \\ \hline
\end{tabular}}%

    \hspace{.5cm}%
  }\hspace{1cm}
  \label{table:3}\subfloat[Experimental results of classifier (Wearing Hat) on dataset (10\% female and 90\% male)]{%
    \resizebox{\linewidth}{!}{\begin{tabular}{|c|cccc|cccc|c|}
\hline
\multirow{2}{*}{methods} & \multicolumn{4}{c|}{simple methods}                                                                                          & \multicolumn{4}{c|}{state-of-the-art methods}                                                                                                 & ours                \\ \cline{2-10} 
                         & \multicolumn{1}{c|}{baseline}   & \multicolumn{1}{c|}{reweighing}  & \multicolumn{1}{c|}{undersampling} & oversampling       & \multicolumn{1}{c|}{ASR}         & \multicolumn{1}{c|}{WFC}                 & \multicolumn{1}{c|}{FAD}                 & FAW                  & AR                  \\ \hline
Accuracy rate (\%)       & \multicolumn{1}{c|}{95.0 (0.5)} & \multicolumn{1}{c|}{93.7 (0.7)}  & \multicolumn{1}{c|}{94.2 (0.8)}    & 94.1 (1.4)         & \multicolumn{1}{c|}{94.5 (0.5)}  & \multicolumn{1}{c|}{92.3 (0.4)}          & \multicolumn{1}{c|}{\textbf{94.6 (0.7)}} & 92.5 (0.7)           & \textbf{95.3 (0.2)} \\ \hline
Disparate Impact (\%)    & \multicolumn{1}{c|}{1.9 (0.6)}  & \multicolumn{1}{c|}{3.3 (0.3)}   & \multicolumn{1}{c|}{2.1 (0.3)}     & \textbf{1.5 (0.3)} & \multicolumn{1}{c|}{2.0 (0.4)}   & \multicolumn{1}{c|}{7.9 (2.6)}           & \multicolumn{1}{c|}{27.2 (1.0)}          & 1.9 (0.5)            & \textbf{0.2 (0.3)}  \\ \hline
Disparate FPR (\%)       & \multicolumn{1}{c|}{-33.0 (1.3)}  & \multicolumn{1}{c|}{-27.2 (5.1)} & \multicolumn{1}{c|}{-29.2 (1.1)}   & -35.5 (2.1)        & \multicolumn{1}{c|}{-23.0 (3.4)} & \multicolumn{1}{c|}{\textbf{17.7 (1.3)}} & \multicolumn{1}{c|}{-30.8 (1.7)}         & \textbf{-18.7 (1.1)} & -22.8 (1.0)         \\ \hline
Disparate FNR (\%)       & \multicolumn{1}{c|}{30.0 (1.2)} & \multicolumn{1}{c|}{36.6 (4.2)}  & \multicolumn{1}{c|}{30.0 (1.3)}    & 29.8 (2.4)         & \multicolumn{1}{c|}{26.8 (0.9)}  & \multicolumn{1}{c|}{\textbf{22.2 (1.8)}} & \multicolumn{1}{c|}{26.9 (1.8)}          & \textbf{26.0 (1.3)}  & 28.2 (2.4)          \\ \hline
\end{tabular}}%

    \hspace{.5cm}%
    
  }
\end{table*}


\begin{figure}
\centering     
\subfloat[male samples assigned with the lowest weights]{\label{fig:a}\includegraphics[width=30mm, height=2.0in]{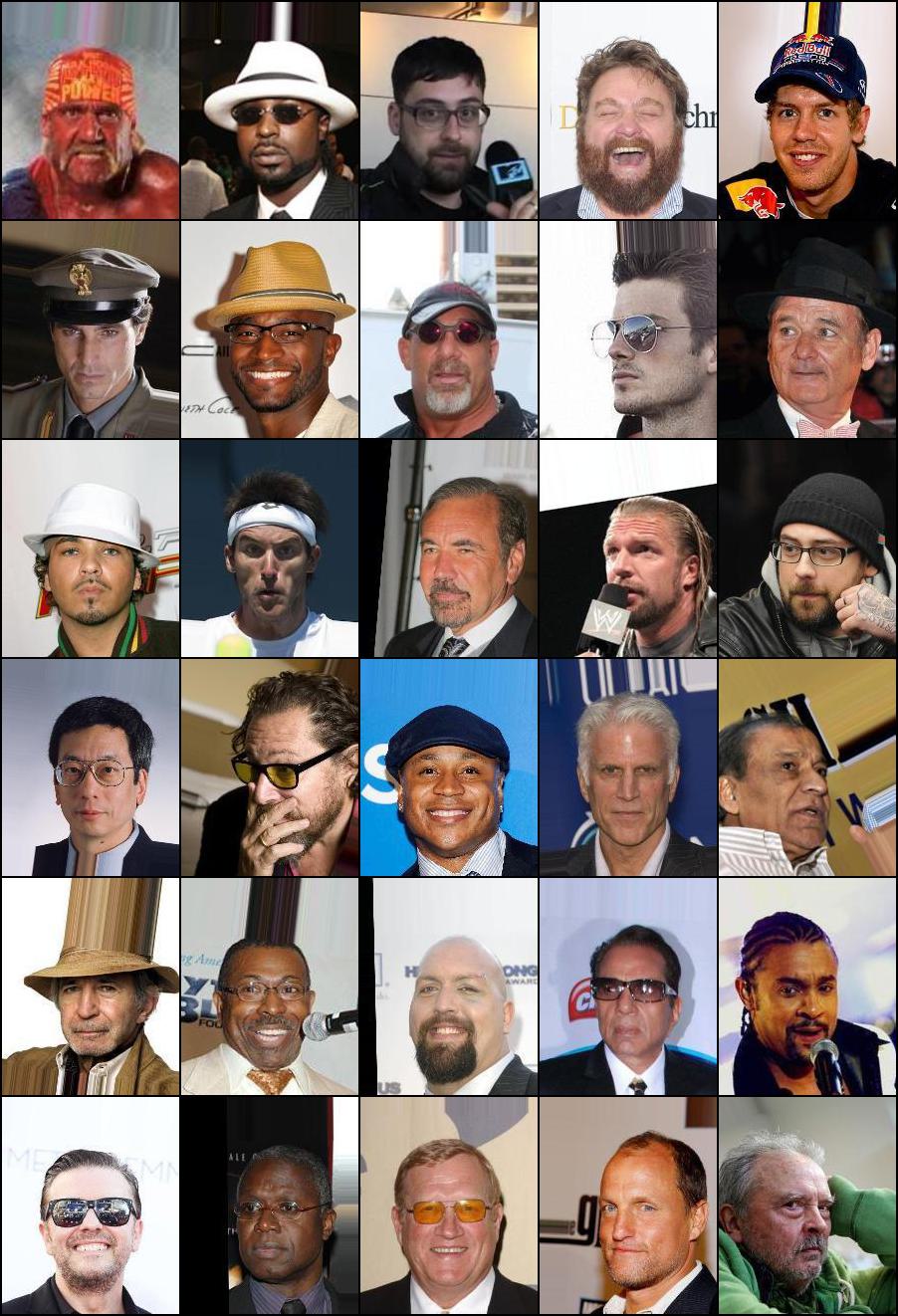}}
\hspace{\floatsep}
\subfloat[male samples assigned with the highest weights]{\label{fig:b}\includegraphics[width=30mm,height=2.0in]{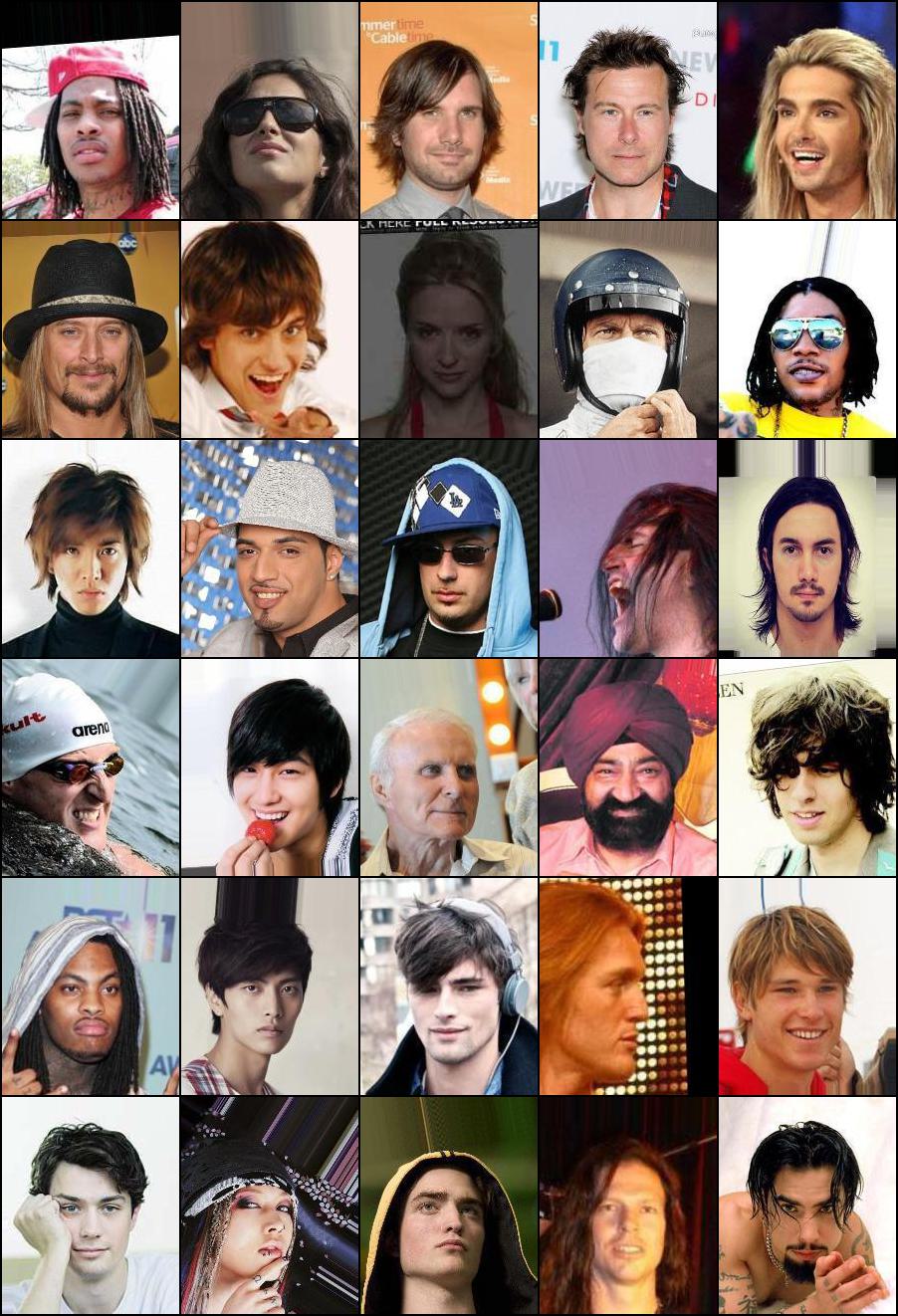}}
\caption{Samples from the male group with the lowest and highest weights. Samples with the lowest weights tend to wear suits and have short hair, while samples with the highest weights tend to have longer hair.}\label{fig:weights}
\end{figure}

\subsubsection{Tabular dataset}
For experiments on tabular data, we use the Adult dataset \citep{kohavi1996} and the UCI German Credit Risk dataset \citep{Dua:2019} (For more details of the datasets, please refer to Appendices). Note that tabular datasets generally need more preprocessing than image datasets~\cite{borisov2022deep}. Note we are aware that gradient boosting would be more adequate for tabular data, but could not find any related approach for mitigating representation bias based on boosting. We normalize the continuous features and use one-hot encoding to deal with the categorical features. We train the model for 50 epochs with
batch-sizes of 1000 and 500 for the male and female samples. For more details of the experiment, please refer to Appendices. 





\subsection{Analysis results}
\subsubsection{Performance comparison}

From Table \ref{table:1} to Table \ref{table:5}, we can see that there is no sacrifice of accuracy with our approach. At the same time, the disparate impact concerning the sensitive attribute is mitigated, which is a crucial advantage of our optimization over related approaches. 
 To better understand the performance of our method, we break down the accuracy concerning the male and female groups and show it in Tables \ref{table:6} and \ref{table:7} in Appendices. 

WFC did not perform so well accuracy-wise. We think it is because it adjusts the prediction results by aligning the Wasserstein distance between the predictions over the sensitive groups, which could reduce the accuracy rate. 
While ASR is a strong competitor, it requires multiple times of training until a convergence of the neural networks is reached, making it more expensive than the other approaches. Similarly, this holds true for FAW. FAD fails to deal with the imbalance problem during the decorrelation, but it maintains a high accuracy rate. 

Figure \ref{fig:weights} shows the samples from the male groups, which are assigned the lowest and highest weights, respectively. We see that 
male samples more distant from the female distribution are down-weighted,  balancing and harmonizing the male and female distributions.
Male samples closer to the female group are assigned relatively high weights, which provides further information for the classification task.



\subsubsection{Embeddings and reweighting visualization}\label{sec:visual}
\begin{figure}
\centering     
\subfloat[t-SNE before reweighting in the latent space]{\label{fig:a1}\includegraphics[width=40mm]{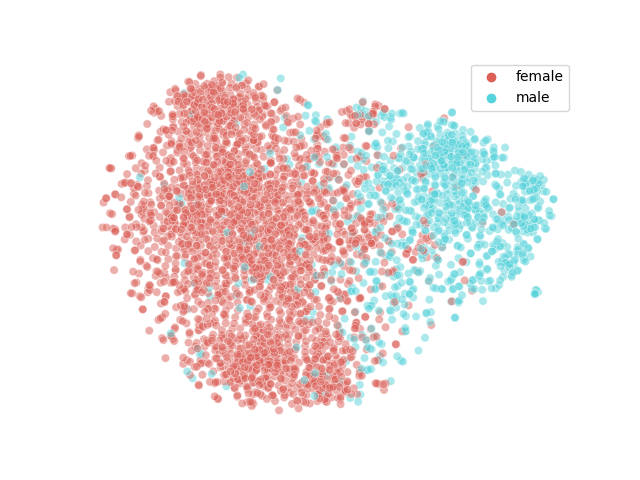}}
\vspace{\floatsep}
\subfloat[t-SNE after reweighting in the latent space]{\label{fig:b1}\includegraphics[width=40mm]{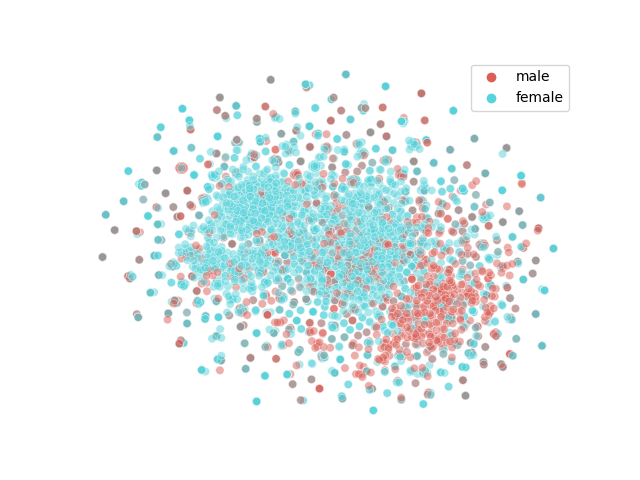}}
\caption{t-SNE of extracted embeddings before and after reweighting of the instances in a setting of 70 \% male and 30 \% female samples}
\label{fig:visual} 
\end{figure}

We visualize the learned weights of the majority group vs.\ the minority group for the 70\% male vs. 30\% female dataset of CelebA. So we show the t-SNE embeddings of the original and reweighted embeddings in Figure \ref{fig:visual}. On the left, in Figure \ref{fig:a1}, we can see that the male and female groups are not aligning well, leading to discrimination against the female group, as described in earlier sections. Our proposed reweighting method aligns the extracted embeddings of the female group to that of the male group, as shown in Figure \ref{fig:b1}, before the classification step. These visualizations, of course, only partially explain our approach's success in dealing with the problem of representation bias concerning a sensitive attribute. In addition, note that the original Wasserstein distance between the two distributions before reweighting is 15.87, and after reweighting, the distance is 0.23. For more details, please refer to Appendices. 

\begin{table*}
\centering
\caption{Experimental results of classifier on Adult dataset (sensitive attribute is gender)}\label{table:4}
\resizebox{\linewidth}{!}{%

\begin{tabular}{|c|cccc|cccc|c|}
\hline
\multirow{2}{*}{methods} & \multicolumn{4}{c|}{simple methods}                                                                                          & \multicolumn{4}{c|}{state-of-the-art methods}                                                                                              & ours                \\ \cline{2-10} 
                         & \multicolumn{1}{c|}{baseline}   & \multicolumn{1}{c|}{reweighing} & \multicolumn{1}{c|}{undersampling} & oversampling        & \multicolumn{1}{c|}{ASR}                & \multicolumn{1}{c|}{WFC}        & \multicolumn{1}{c|}{FAD}                 & FAW                 & AR                  \\ \hline
Accuracy rate (\%)       & \multicolumn{1}{c|}{83.1 (0.4)} & \multicolumn{1}{c|}{82.5 (0.3)} & \multicolumn{1}{c|}{82.1 (0.3)}    & \textbf{84.7 (0.9)} & \multicolumn{1}{c|}{81.6 (0.3)}         & \multicolumn{1}{c|}{81.8 (0.5)} & \multicolumn{1}{c|}{82.4 (0.5)}          & 81.2 (0.6)          & \textbf{83.0 (0.1)} \\ \hline
Disparate Impact (\%)    & \multicolumn{1}{c|}{17.8 (0.3)} & \multicolumn{1}{c|}{21.0 (0.4)} & \multicolumn{1}{c|}{18.7 (0.5)}    & 18.6 (0.4)          & \multicolumn{1}{c|}{\textbf{0.4 (0.2)}} & \multicolumn{1}{c|}{2.5 (1.0)}  & \multicolumn{1}{c|}{5.7 (1.4)}           & 1.7 (0.4)           & \textbf{1.3 (0.5)}  \\ \hline
Disparate FPR (\%)       & \multicolumn{1}{c|}{17.0 (1.0)} & \multicolumn{1}{c|}{\textbf{2.3 (4.7)}}  & \multicolumn{1}{c|}{9.2 (1.3)}     & 8.4 (1.6)           & \multicolumn{1}{c|}{27.2 (4.5)}         & \multicolumn{1}{c|}{-9.8 (0.7)} & \multicolumn{1}{c|}{\textbf{-8.7 (1.8)}} & -8.5 (2.4) & -10.5 (1.1)         \\ \hline
Disparate FNR (\%)       & \multicolumn{1}{c|}{6.1 (0.6)}  & \multicolumn{1}{c|}{12.1 (3.5)} & \multicolumn{1}{c|}{4.2 (0.8)}     & 12.7 (0.7)          & \multicolumn{1}{c|}{\textbf{2.3 (1.2)}} & \multicolumn{1}{c|}{22.4 (1.3)} & \multicolumn{1}{c|}{\textbf{3.2 (0.7)}}  & 4.0 (1.7)           & 7.2 (0.7)           \\ \hline
\end{tabular}}
\end{table*}
\begin{table*}
\centering
\caption{Experimental results of the classifier on German Credit dataset (sensitive attribute is sex)}\label{table:5}
\resizebox{\linewidth}{!}{%

\begin{tabular}{|c|cccc|cccc|c|}
\hline
\multirow{2}{*}{methods} & \multicolumn{4}{c|}{simple methods}                                                                                   & \multicolumn{4}{c|}{state-of-the-art methods}                                                                                              & ours               \\ \cline{2-10} 
                         & \multicolumn{1}{c|}{baseline}   & \multicolumn{1}{c|}{reweighing} & \multicolumn{1}{c|}{undersampling} & oversampling & \multicolumn{1}{c|}{ASR}                 & \multicolumn{1}{c|}{WFC}                & \multicolumn{1}{c|}{FAD}                 & FAW        & AR                 \\ \hline
Accuracy rate (\%)       & \multicolumn{1}{c|}{70.1 (1.4)} & \multicolumn{1}{c|}{69.2 (1.3)} & \multicolumn{1}{c|}{65.5 (1.1)}    & 67.3 (2.9)   & \multicolumn{1}{c|}{69.1 (2.4)} & \multicolumn{1}{c|}{\textbf{69.7 (0.5)} }         & \multicolumn{1}{c|}{68.9 (0.5)} & 69.5 (0.5) & \textbf{70.0 (1.4)}         \\ \hline
Disparate Impact (\%)    & \multicolumn{1}{c|}{15.4 (9.3)} & \multicolumn{1}{c|}{8.9 (3.4)}  & \multicolumn{1}{c|}{11.9 (6.5)}    & 9.2 (4.7)    & \multicolumn{1}{c|}{6.7 (2.5)}           & \multicolumn{1}{c|}{\textbf{3.2 (1.0)}} & \multicolumn{1}{c|}{3.5 (0.4)}           & 3.8 (0.5)  & \textbf{1.1 (1.2)} \\ \hline
Disparate FPR (\%)       & \multicolumn{1}{c|}{7.5 (4.8)}  & \multicolumn{1}{c|}{19.0 (3.7)} & \multicolumn{1}{c|}{13.4 (7.3)}    & 12.5 (8.5)   & \multicolumn{1}{c|}{\textbf{1.5 (0.9)}}  & \multicolumn{1}{c|}{7.2 (0.8)}          & \multicolumn{1}{c|}{\textbf{1.5 (0.2)}}  & 9.2 (2.8)  & 5.3 (1.7)          \\ \hline
Disparate FNR (\%)       & \multicolumn{1}{c|}{6.3 (3.2)}  & \multicolumn{1}{c|}{9.0 (3.2)}  & \multicolumn{1}{c|}{15.7 (8.8)}    & 11.8 (8.7)   & \multicolumn{1}{c|}{\textbf{0.9 (0.4)}}  & \multicolumn{1}{c|}{5.2 (2.3)}          & \multicolumn{1}{c|}{\textbf{1.1 (0.3)}}  & 4.9 (2.5)  & 6.7 (0.9)          \\ \hline
\end{tabular}}
\end{table*}

\subsubsection{Classification with noisy label}
\begin{figure}
\centering     
\subfloat[Accuracy vs. Noise ratio]{\label{fig:noise1}\includegraphics[width=30mm,height=1.0in]{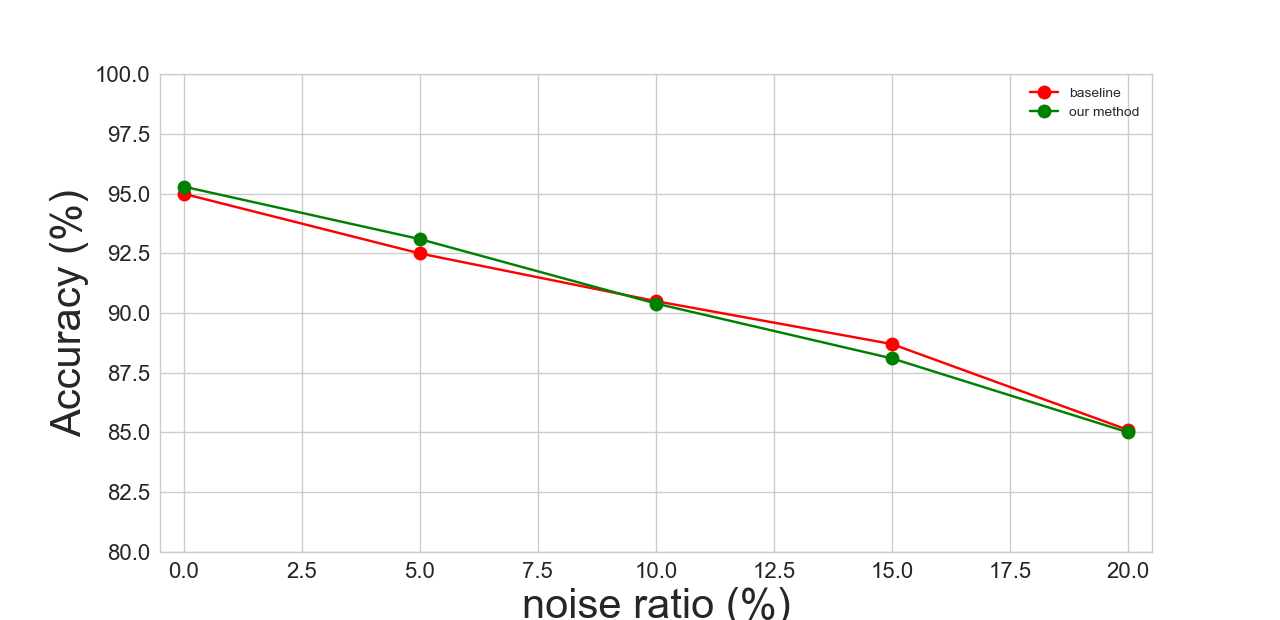}}
\hspace{\floatsep}
\subfloat[Disparate Impact vs. Noise ratio]{\label{fig:noise2}\includegraphics[width=30mm,height=1.0in]{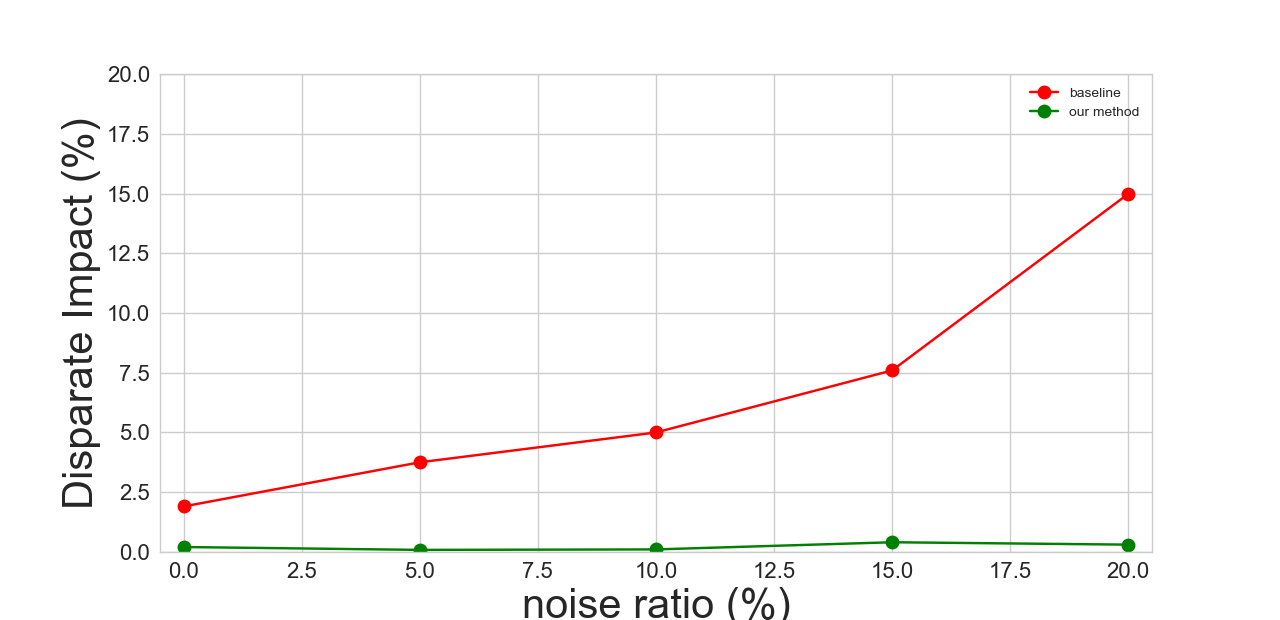}}
\caption{Change of accuracy and disparate impact under different noise ratios on CelebA 90\% male and 10\% female. Note that we enforce Demographic Parity which is selection rate parity 
(which our method can implicitly mitigate) and we believe this is the reason why our method is so robust to noisy labeling.} 
 \label{fig:noise}
\end{figure}

Since our method ensures the demographic parity of the predictions, it should not be sensitive to noise labeling (possible biased labeling). We apply half the noise
corruption to the majority group and half to the minority group. We show the performance of our method and baseline (NN-based classification without fairness constraints) on accuracy and disparate impact under different ratios of noise corruption in Figure \ref{fig:noise}. The disparate impact remains low when the noise ratio changes. Moreover, again, we show no sacrifice of accuracy when applying our method. For more details of performance on other fairness metrics, please refer to Figure \ref{fig:noise_} in Appendices.

\subsubsection{Sensitivity to the choice of hyper-parameters}
We have also analyzed the sensitivity of our method to the hyper-parameter $T$ mentioned earlier, in Figure \ref{fig:sens1} in Appendices, where the plots indicate that the performance of our adversarial reweighting scheme has low sensitivity to the choice of the hyper-parameter. In our experiments, we set $T$ at 5. For analysis of other datasets, see Figure \ref{fig:sens2} in Appendices. 

\subsubsection{Ablation tests} 
$ $

\noindent
\textit{\textbf{Ablation test for MMD and JS-divergence dissimilarity measures.}}
We also conducted an ablation test for the
Jenson-Shannon-divergence (JS) and maximum-mean discrepancy (MMD) instead of Wasserstein distance to learn the weights in our framework on the CelebA dataset, with 90\% male and 10\% female samples. In Figures \ref{fig:abalation1} and \ref{fig:ablation2} (see Appendices for Figure \ref{fig:ablation2}), the performance of our method using the Wasserstein distance is better than JS and MMD. Wasserstein distance may be more suitable to measure their distance than the JS divergence when the distributions are more disjoint. MMD with kernels may be unable to capture very complex distances in high dimensional spaces compared to Wasserstein distance. The Wasserstein distance is better for accuracy and disparate impact but not necessarily better at Disparate FPR and Disparate FNR.

\begin{figure}
\centering     
\subfloat[Accuracy vs. Distance Methods]{\label{fig:a4}\includegraphics[width=30mm,height=0.8in]{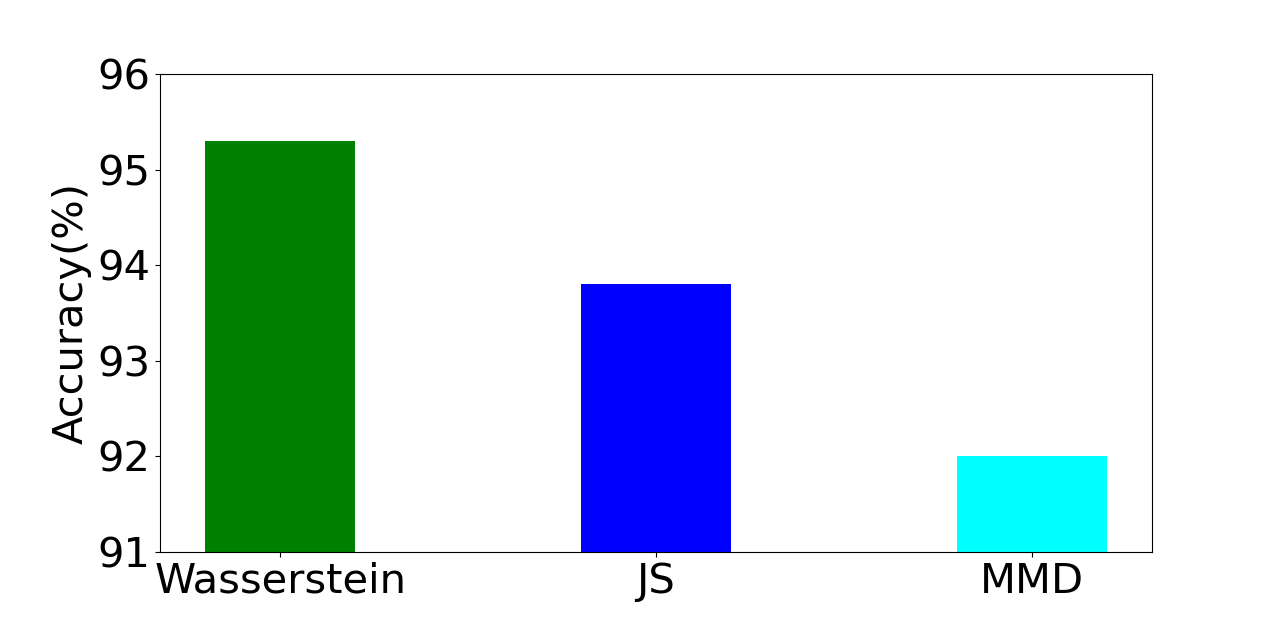}}
\hspace{\floatsep}
\subfloat[Disparate Impact vs. Distance Methods]{\label{fig:b4}\includegraphics[width=30mm,height=0.8in]{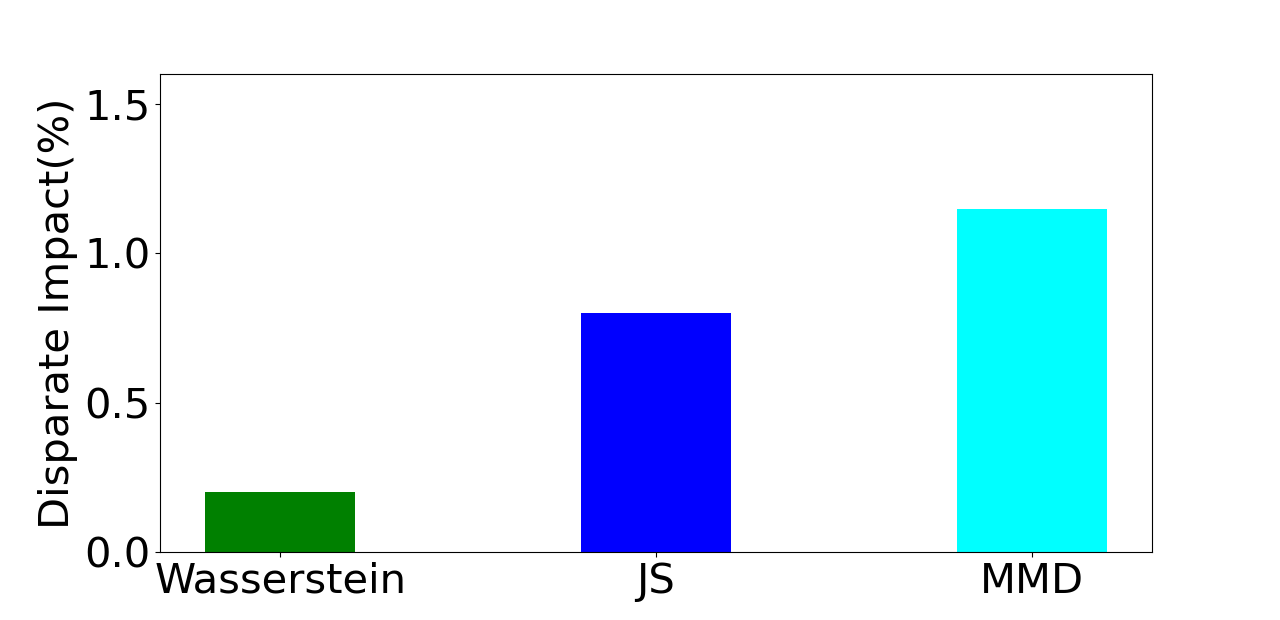}}

\subfloat[Disparate FPR vs. Distance Methods]
{\label{fig:c43}\includegraphics[width=30mm,height=0.8in]{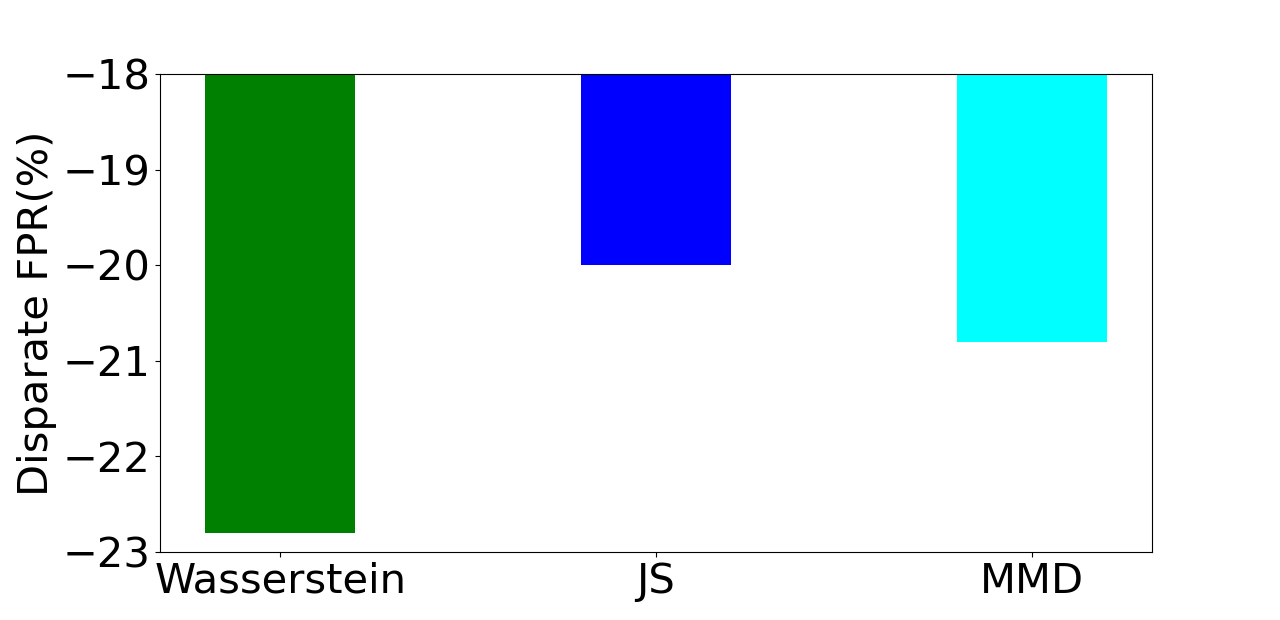}}
\hspace{\floatsep}
\subfloat[Disparate FNR vs. Distance Methods]{\label{fig:d4}\includegraphics[width=30mm,height=.8in]{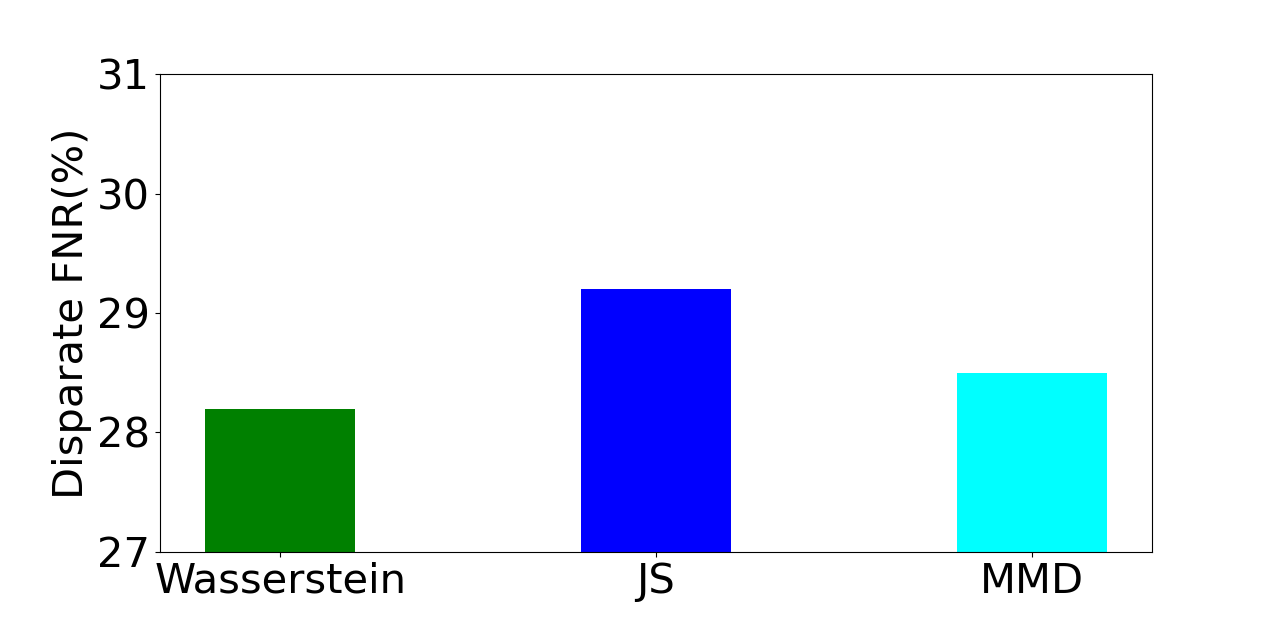}}
\caption{Sensitivity of different distance measures on CelebA dataset with 90\% male and 10\% female}\label{fig:abalation1}
\end{figure}

\noindent
\textit{\textbf{Ablation test for assigning weights to both groups or only to the minority group.}}
Our method, by design, aims at reweighting the majority group to close the gap to the Wasserstein distance between samples with different sensitivity attribute values. Therefore, we test various reweighting schemes for both groups in an ablation test by alternatively assigning weights to one while treating another group with fixed weights -- the Wasserstein distance between the two groups is 0.17 for the dataset 70\% male and 30\% female images from CelebA. We also test the assignment of weights to the minority group. In this situation, the Wasserstein distance after reweighting is 1.29. As we mentioned earlier, the Wasserstein distance of our method after reweighting is 0.23. This suggests that the difference between reweighting both and only the majority group is marginal, while the difference between reweighting only the majority and the minority group is significant. This is why we reweight the majority group in our method. For more details, please refer to Appendices. 

\section{Discussion and Conclusion}\label{sec:discussion}

Our work is conceptually different from previous works on fairness-aware machine learning because
it balances and harmonizes protected groups, as defined by sensitive attributes, minimizes the empirical risk, and 
achieves competitive predictive quality regarding accuracy, fairness, and robustness. Theoretical literature explores the inherent balance between fairness and utility, and numerous experiments have demonstrated the trade-off in practice. 
Nevertheless, these trade-off discussions are often based on a fixed distribution that does not align with our current situation. We argue that an ideal distribution exists where fairness and utility are in harmony. Our data reweighing combined with classifier training lets us move beyond a biased distribution and release the trade-off. One limitation in our work lies in that WGAN-GP might fail to approximate the Wasssertein distance correctly \cite{Mallasto2019, Stanczuk2021}. Based on our experiments, our method can mitigate the need to discard sensitive attributes or impose specific fairness constraints, thus avoiding the issue of determining critical hyperparameters such as regularization factors. Our approach still permits the inclusion of further regularizations or constraints during empirical risk minimization --- though we have not yet found the need to explore such approaches. 

\bibliography{aaai24}


\appendix


\section{Appendices}

\subsection{Mathematical Details}
We prove that enforcing the Wasserstein distance, indicated as $W(\cdot, \cdot)$, being small in the latent space enforces it to be small in the prediction space as well. 
\begin{proof}
    The proof goes as follows\footnote{The proof is almost verbatim the comment from the user Christian Bueno about statistical divergence change under a Lipschitz push-forward map at \url{https://mathoverflow.net/questions/314201/how-does-a-statistical-divergence-change-under-a-lipschitz-push-forward-map}.}:

\begin{subequations}\label{proof}
   \begin{align*}
      W(C_{\#}\mu, C_{\#}\nu) & = \sup_{f\in Lip_1(Y)}\: \int_Y fdC_{\#}\mu - \int_Y fdC_{\#}\nu \\&\text{(Kantorovich duality)}\\
                              & = \sup_{f\in Lip_1(Y)}\: \int_Z f \circ C d\mu - \int_Z f \circ C d\nu \\&\text{(Property of the push-forward)}\\
                              & = \sup_{f\in Lip_1(Y)}\: K\cdot\left(\int_Z \frac{f \circ C}{K} d\mu - \int_Z \frac{f \circ C}{K} d\nu\right) & & \\
                            & \leq \sup_{h\in Lip_1(Z)}\: K \cdot\left(\int_Z h d\mu - \int_Z h  d\nu\right) \\&\text{(}\frac{f \circ C}{K}\text{ is 1-Lipschitz)}\\
                               & =  K\cdot W(\mu, \nu) &\tag{4} 
    \end{align*}
    \end{subequations}

    Where $Lip_1(Y)$ indicates the set of 1-Lipschitz functions $f:Y \to \mathbb{R}$.
\end{proof}

\subsection{Dataset Details}

\subsubsection{CelebA}\label{celeba} CelebA contains 202,600 face images, each endowed with 40 attributes. When we try to construct the datasets from CelebA based on our needs, we maintain the class imbalance in the three datasets constant, which is 70\% not wearing hats and 30\% wearing hats, since class imbalance is not our priority in this paper.

\subsubsection{Adult dataset}\label{adult} The Adult dataset was drawn from the 1994 United States Census Bureau data. It used personal information such as education level and working hours per week to predict whether an individual earns more or less than \$50,000 per year. The dataset is imbalanced -- the instances made less than \$50,000 constitute 25\% of the dataset, and the instances made more than \$50,000 include 75\% of the dataset. As for gender, it is also imbalanced. We use age, years of education, capital gain, capital loss, hours-per-week, etc., as continuous features, and education level, gender, etc., as categorical features.

\subsubsection{UCI German Credit Risk dataset}\label{lending}
This dataset contains 1000 entries with 20 categorial/symbolic attributes. In this dataset, each entry represents a person who takes credit from a bank. Each person is classified as having good or bad credit risks according to their attributes.

\subsection{Training Details and Results}\label{training}

\subsubsection{Details of WGAN-GP adaptation for our method} In the original design of WGAN-GP of the training for one batch, the sizes of generated and original samples are equal for the Gradient Penalty as a regularizer to be applied. Here we need to make some changes: we control the sum of the majority group by the weights in one batch by the Wasserstein distance to let it be equal to the sample size of the minority group in one batch. Then we send them for further computation of the regularizer.

\subsubsection{Repetition} We repeat experiments on each dataset five times. Before each repetition, we randomly split data into training data and test data for the computation of the standard errors of the metrics.

\subsubsection{CelebA training}\label{sec:training-celeba}
\begin{figure}
\centering     
\subfloat[Disparate FPR vs. Noise ratio]{\label{fig:noise3}\includegraphics[width=40mm]{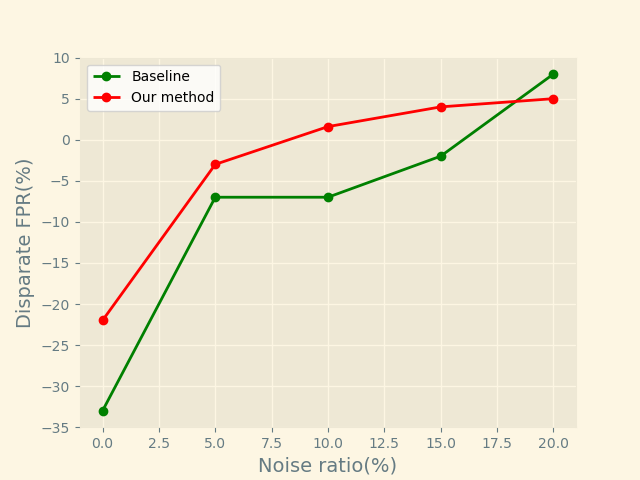}}
\vspace{\floatsep}
\subfloat[Disparate FNR vs. Noise ratio]{\label{fig:noise4}\includegraphics[width=40mm]{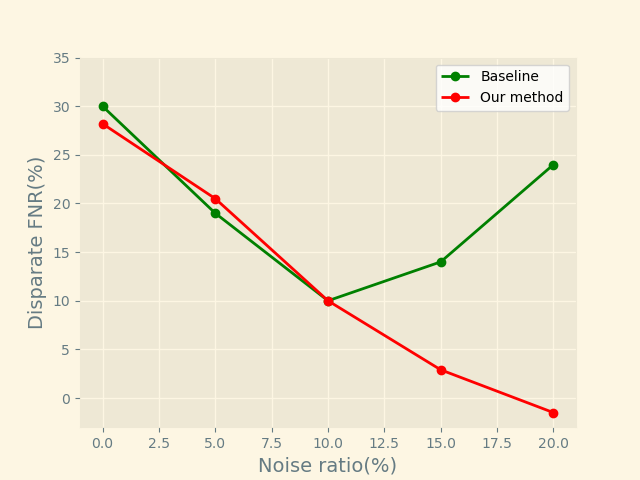}}
\caption{Change of disparate FPR and disparate FNR under different noise ratios on CelebA 90\% male and 10\% female. 
} \label{fig:noise_}
\end{figure}

For the CelebA dataset, since the original data is highly dimensional image data, we use ResNet18 and remove the last layer as a feature extractor. The dimension of the latent space is 512. Note that we use relatively large batch sizes during the training, and we control the sizes of the majority and minority constant during each batch. Papers mention that the large batch size could cause the potential failure of the approximation using neural networks to evaluate the distributions. Our training dataset has 10000 samples, and the test dataset has 2000 samples for all three datasets. From Figure \ref{fig:noise_}, we can see that our method has its limitation regarding Disparate FPR and Disparate DNR. 

\subsubsection{Convergence of the training loss}
\begin{figure}
    \centering
    \includegraphics[width=.3\textwidth]{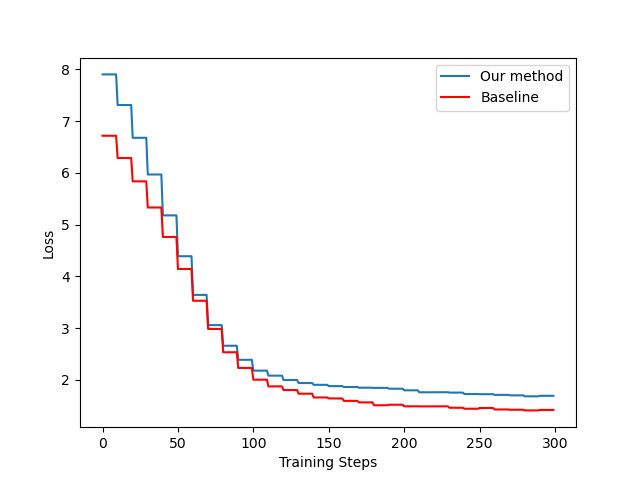}
    \caption{convergence of training loss
}

    \label{fig:convergence}
\end{figure}

We try to show the stability of training of our method. Figure \ref{fig:convergence} shows our method’s and baseline’s convergence for the CelebA dataset with 90\% male and 10\% female.

\subsubsection{Details to approximate Wasserstein distance with and without reweighting}\label{w_reweighting} To approximate the Wasserstein distance before reweighting, we train the feature extractor and discriminator without the weights assigned to any samples. When the NNs are trained, we use the discriminator to approximate the Wasserstein distance between the two groups. 

\subsubsection{Breakdown of accuracy on sensitive groups} \label{sec:breakdown} We could see that the method sacrifices the accuracy of the majority group for the accuracy of the minority group in Table \ref{table:6} and \ref{table:7}.

\begin{table}[ht]
\begin{tabular}{cllll}
\cline{1-4}
\multicolumn{1}{|c|}{\multirow{2}{*}{methods}} & \multicolumn{3}{c|}{accuracy (\%)}                                                                     &  \\ \cline{2-4}
\multicolumn{1}{|c|}{}                         & \multicolumn{1}{l|}{male group} & \multicolumn{1}{l|}{female group} & \multicolumn{1}{c|}{total}  &  \\ \cline{1-4}
\multicolumn{1}{|c|}{baseline}                 & \multicolumn{1}{l|}{95.2}     & \multicolumn{1}{l|}{94.8}       & \multicolumn{1}{l|}{95.1}  &  \\ \cline{1-4}
\multicolumn{1}{|l|}{our method}               & \multicolumn{1}{l|}{94.6}       & \multicolumn{1}{l|}{94.9}       & \multicolumn{1}{l|}{94.7} &  \\ \cline{1-4}
\multicolumn{1}{l}{}                           &                                 &                                   &                             & 
\end{tabular}

\caption{breakdown of accuracy on 70\% male and 30\% female CelebA dataset}\label{table:6}
\end{table}

\begin{table}[ht]
\begin{tabular}{cllll}
\cline{1-4}
\multicolumn{1}{|c|}{\multirow{2}{*}{methods}} & \multicolumn{3}{c|}{accuracy (\%)}                                                                     &  \\ \cline{2-4}
\multicolumn{1}{|c|}{}                         & \multicolumn{1}{l|}{male group} & \multicolumn{1}{l|}{female group} & \multicolumn{1}{c|}{total}  &  \\ \cline{1-4}
\multicolumn{1}{|c|}{baseline}                 & \multicolumn{1}{l|}{95.6}     & \multicolumn{1}{l|}{92.6}       & \multicolumn{1}{l|}{95.0}  &  \\ \cline{1-4}
\multicolumn{1}{|l|}{our method}               & \multicolumn{1}{l|}{95.1}       & \multicolumn{1}{l|}{94.1}       & \multicolumn{1}{l|}{95.3} &  \\ \cline{1-4}
\multicolumn{1}{l}{}                           &                                 &                                   &                             & 
\end{tabular}
\caption{breakdown of accuracy on 90\% male and 10\% female CelebA dataset}\label{table:7}
\end{table}

\begin{figure}
\centering     
\subfloat[Accuracy vs. Distance Methods]{\label{fig:a41}\includegraphics[width=35mm,height=1.0in]{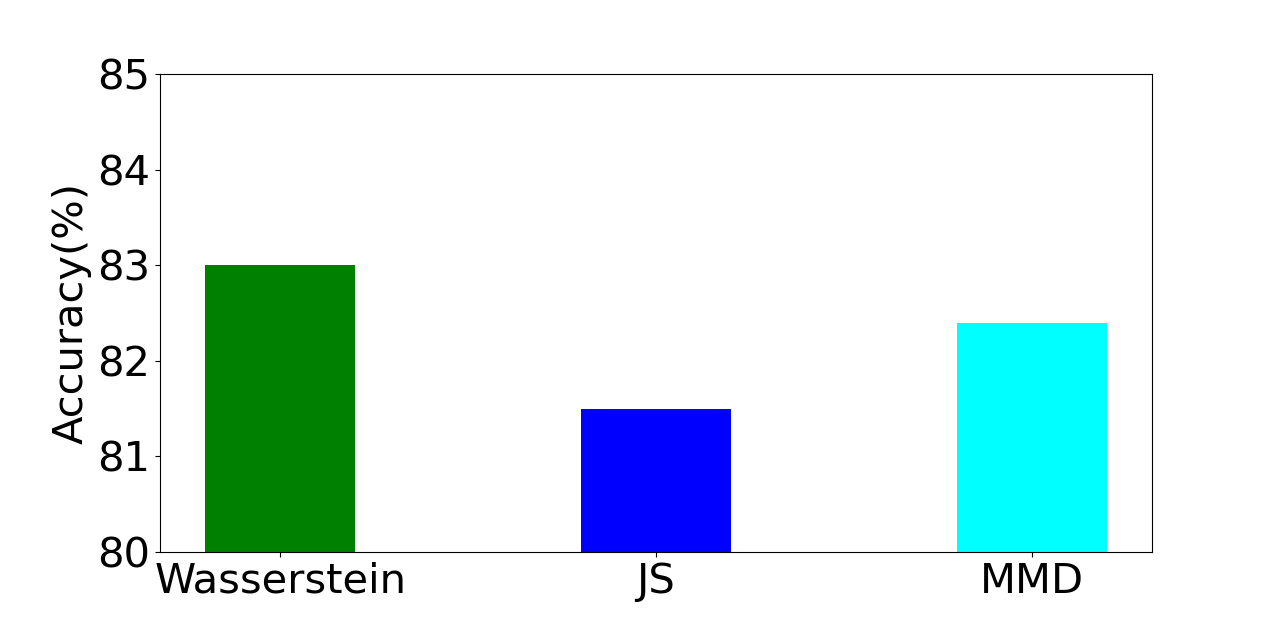}}
\hspace{\floatsep}
\subfloat[Disparate Impact vs. Distance Methods]{\label{fig:b41}\includegraphics[width=35mm,height=1.0in]{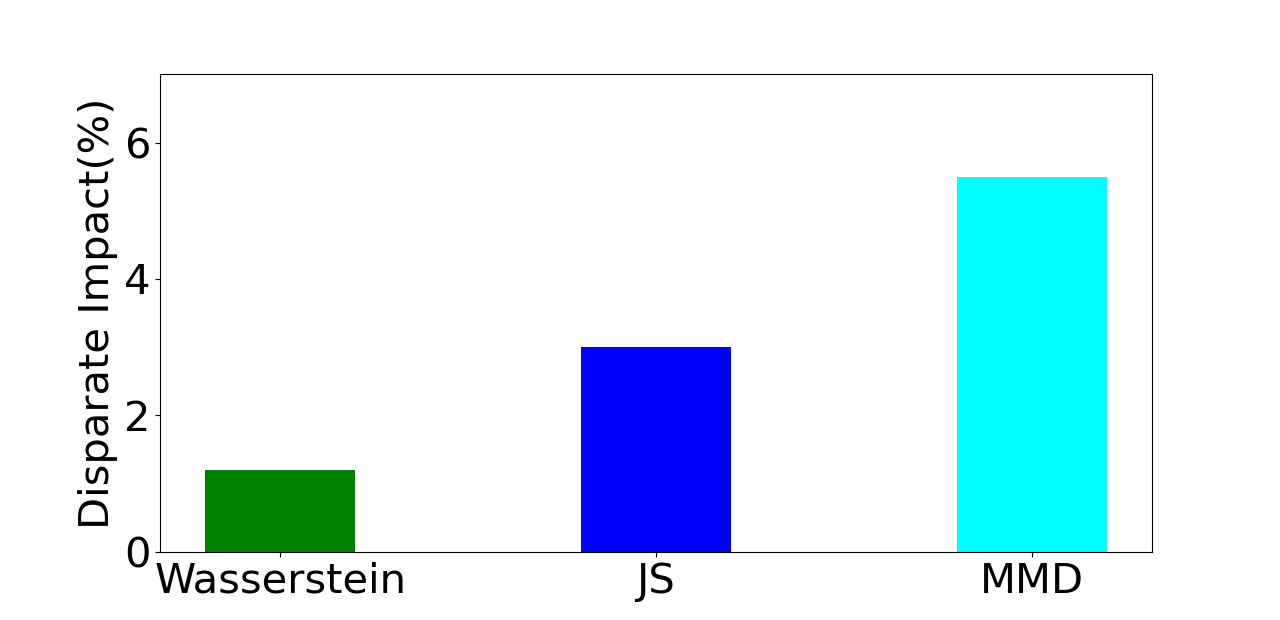}}

\subfloat[Disparate FPR vs. Distance Methods]
{\label{fig:c41}\includegraphics[width=35mm,height=1.0in]{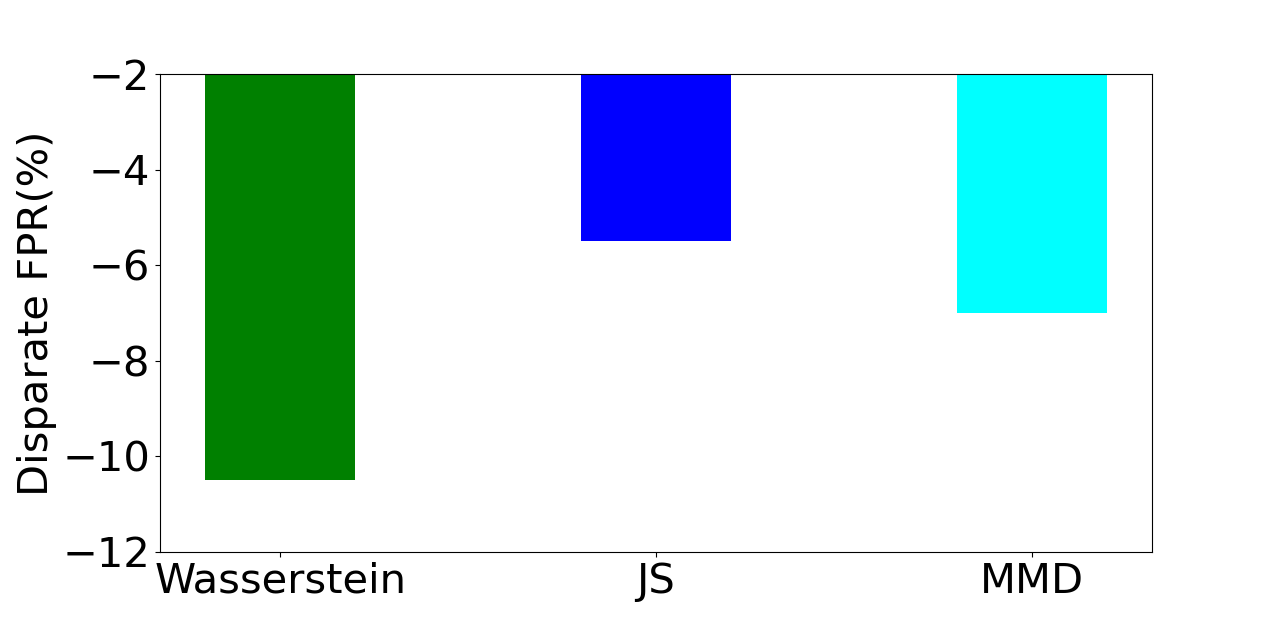}}
\hspace{\floatsep}
\subfloat[Disparate FNR vs. Distance Methods]{\label{fig:d41}\includegraphics[width=35mm,height=1.0in]{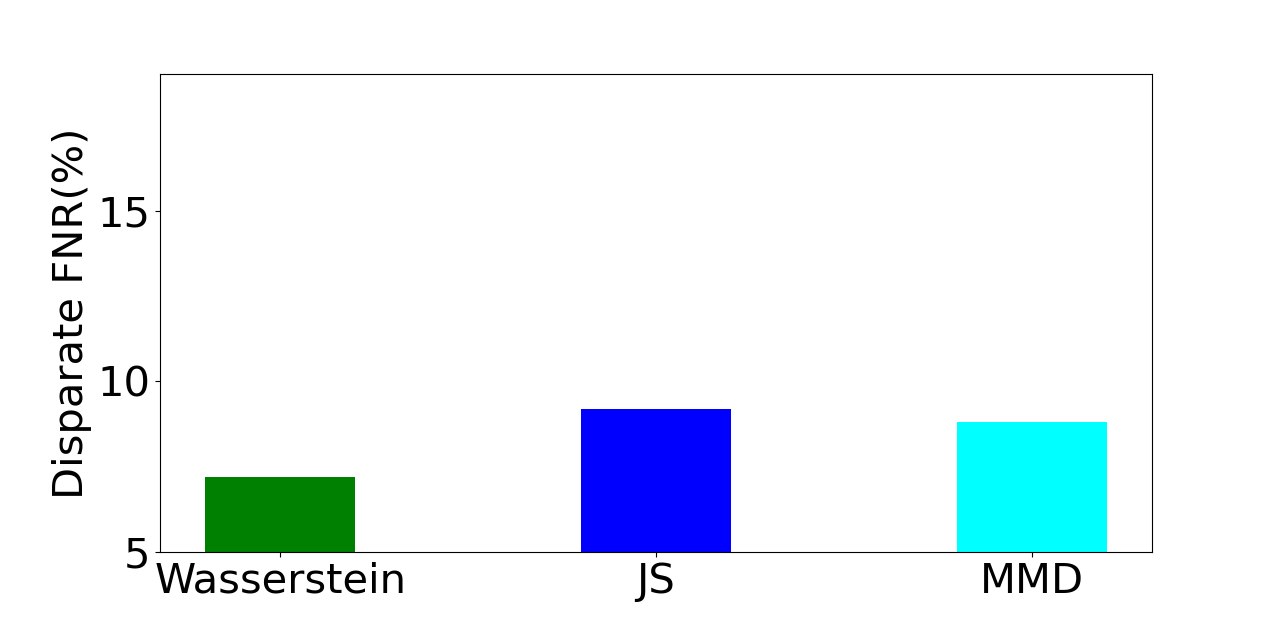}}
\caption{sensitivity of different distance measures on Adult dataset}\label{fig:ablation2}
\end{figure}

\vspace{0.00mm}

\begin{figure}[!t]
\centering     
\subfloat[Accuracy vs. T]{\label{fig:a2}\includegraphics[width=40mm,height=1.2in]{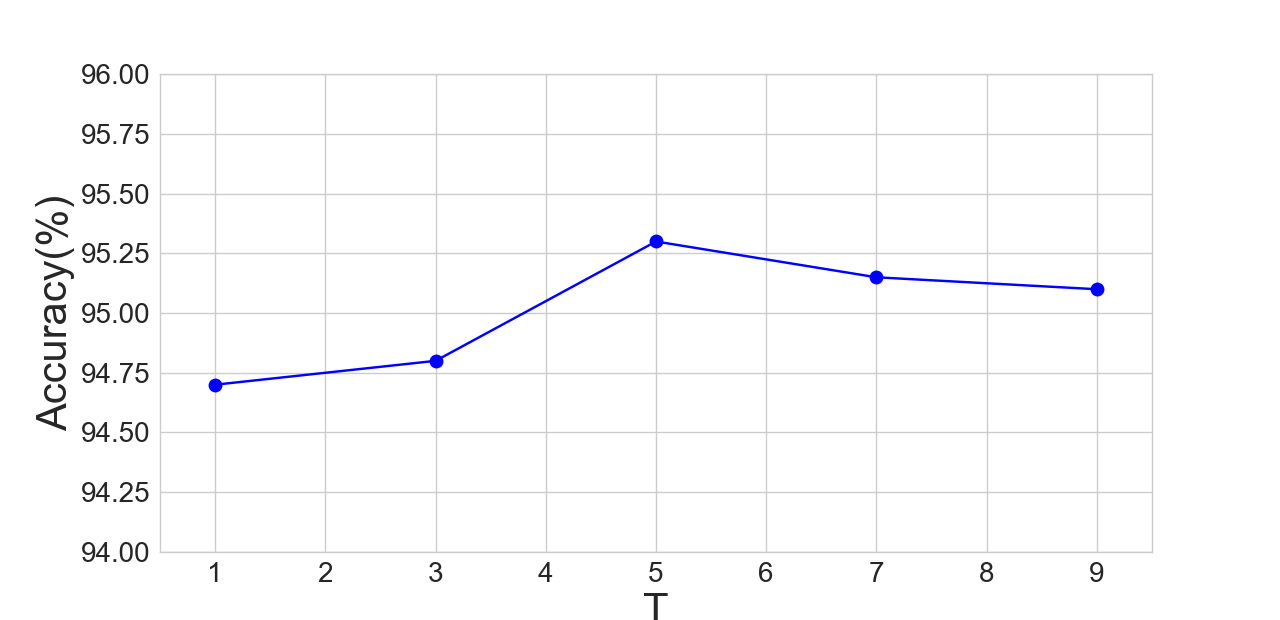}}
\subfloat[Disparate Impact vs. T]{\label{fig:b2}\includegraphics[width=40mm,height=1.2in]{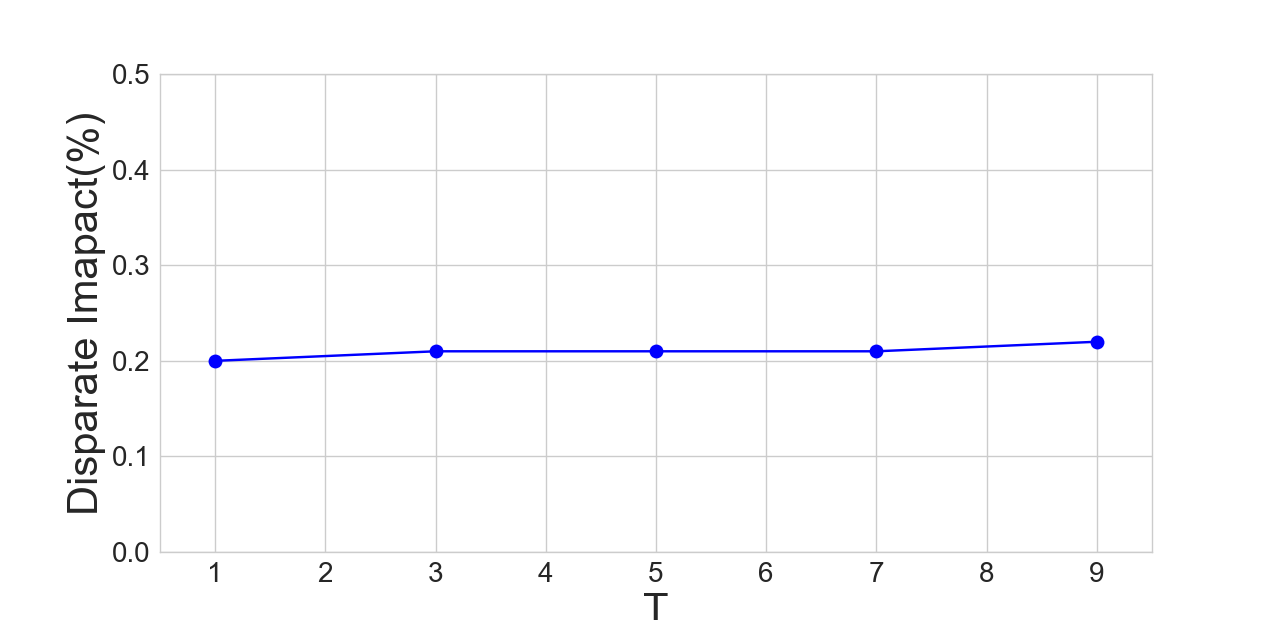}}

\subfloat[Disparate FPR vs. T]
{\label{fig:c2}\includegraphics[width=40mm,height=1.2in]{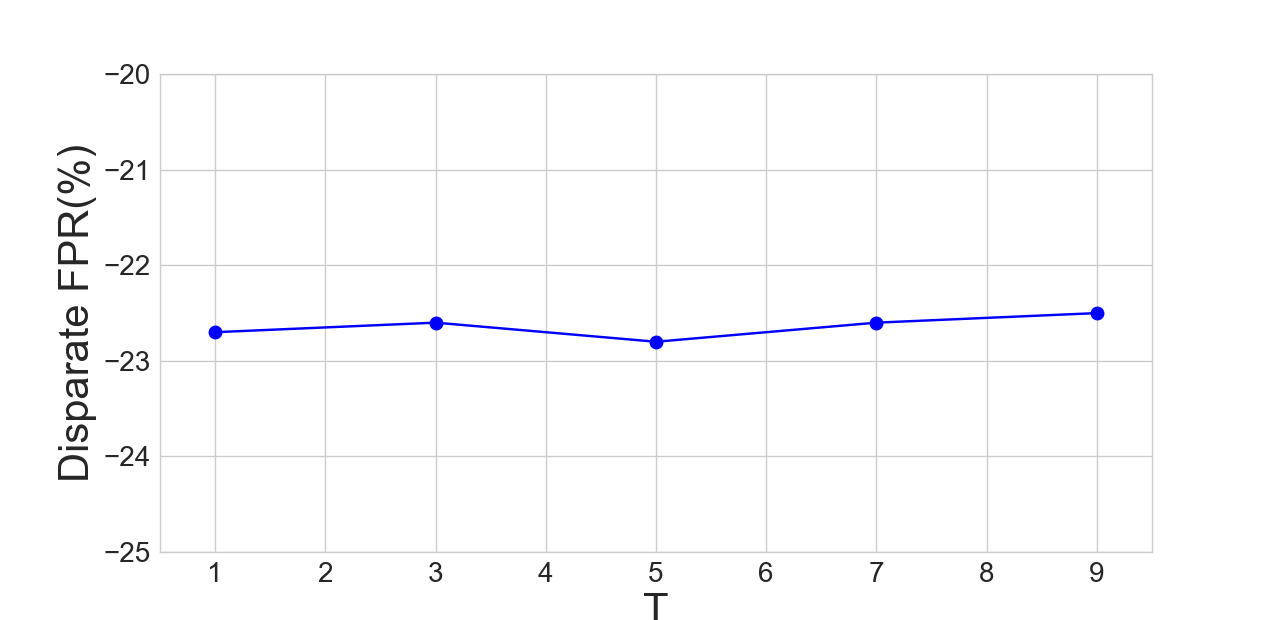}}
\subfloat[Disparate FNR vs. T]{\label{fig:d2}\includegraphics[width=40mm,height=1.2in]{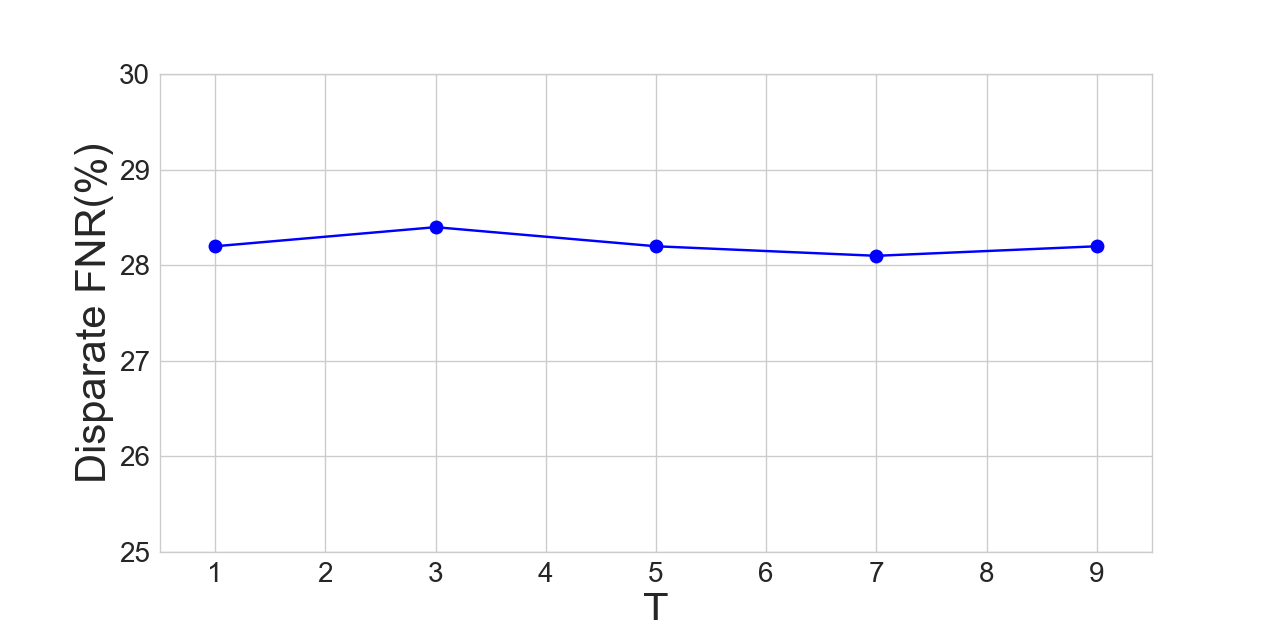}}
\caption{Sensitivity of metrics on the change of $T$ on CelebA 90\% male and 10\% female. 
} \label{fig:sens1}
\end{figure}

\vspace{0.00mm}

\begin{figure}[!t]
\centering     
\subfloat[Accuracy vs. T]{\label{fig:a3}\includegraphics[width=40mm,height=1.2in]{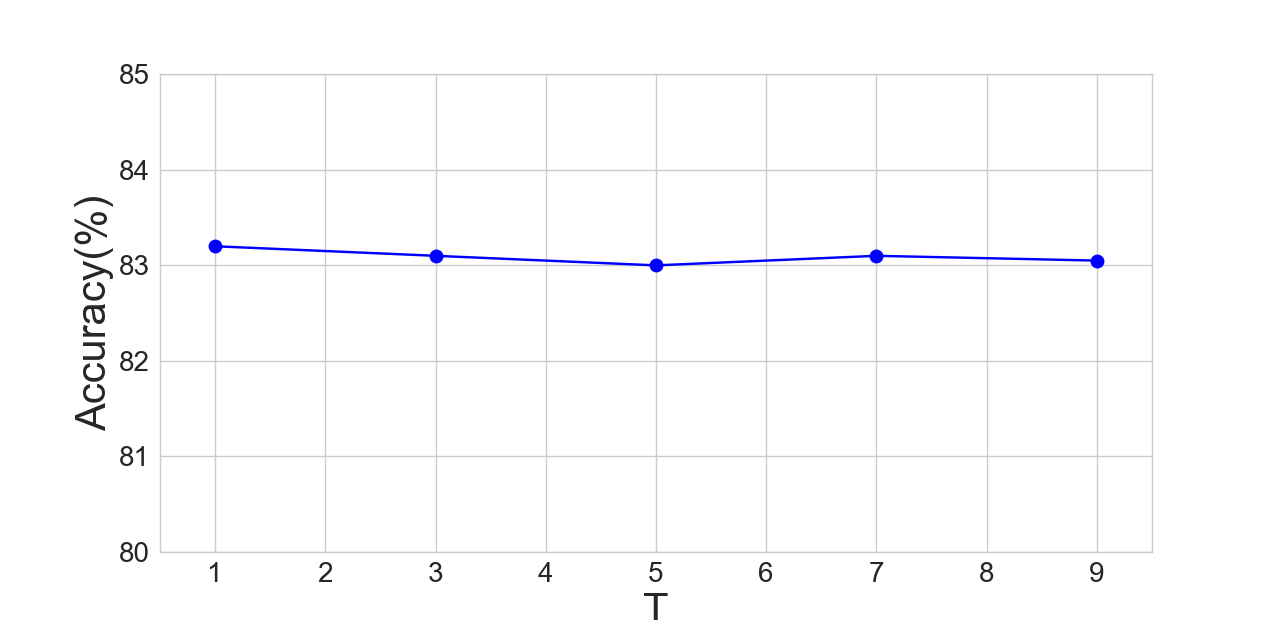}}
\subfloat[Disparate Impact vs. T]{\label{fig:b3}\includegraphics[width=40mm,height=1.2in]{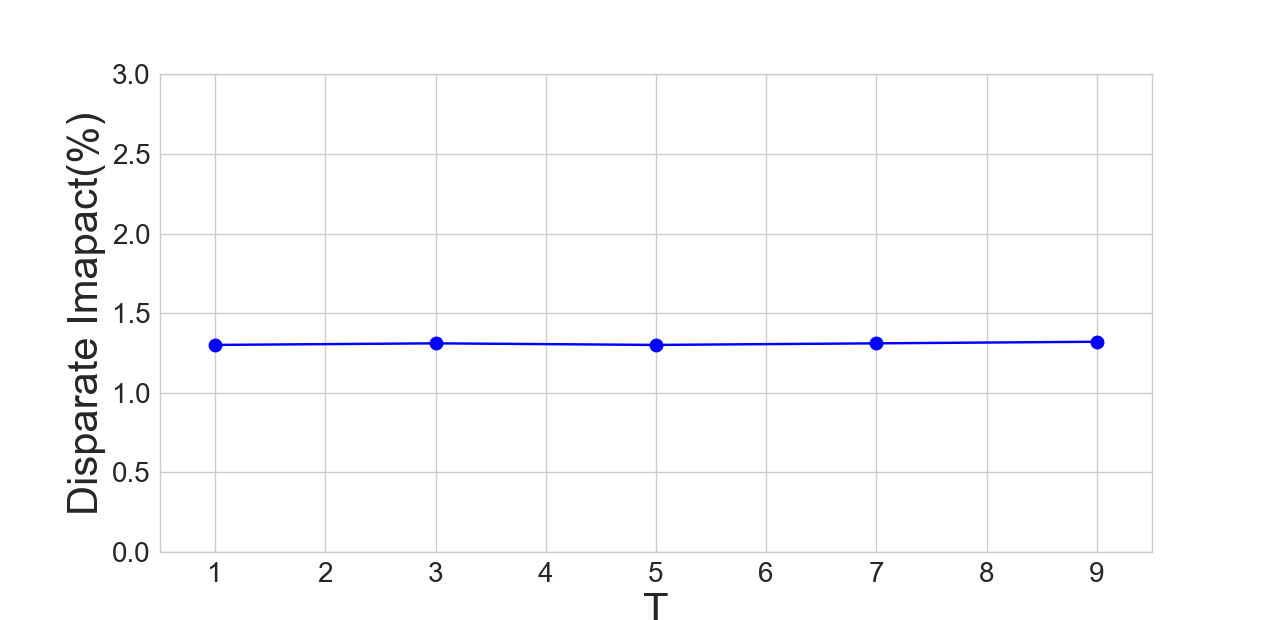}}

\subfloat[Disparate FPR vs. T]
{\label{fig:c3}\includegraphics[width=40mm,height=1.2in]{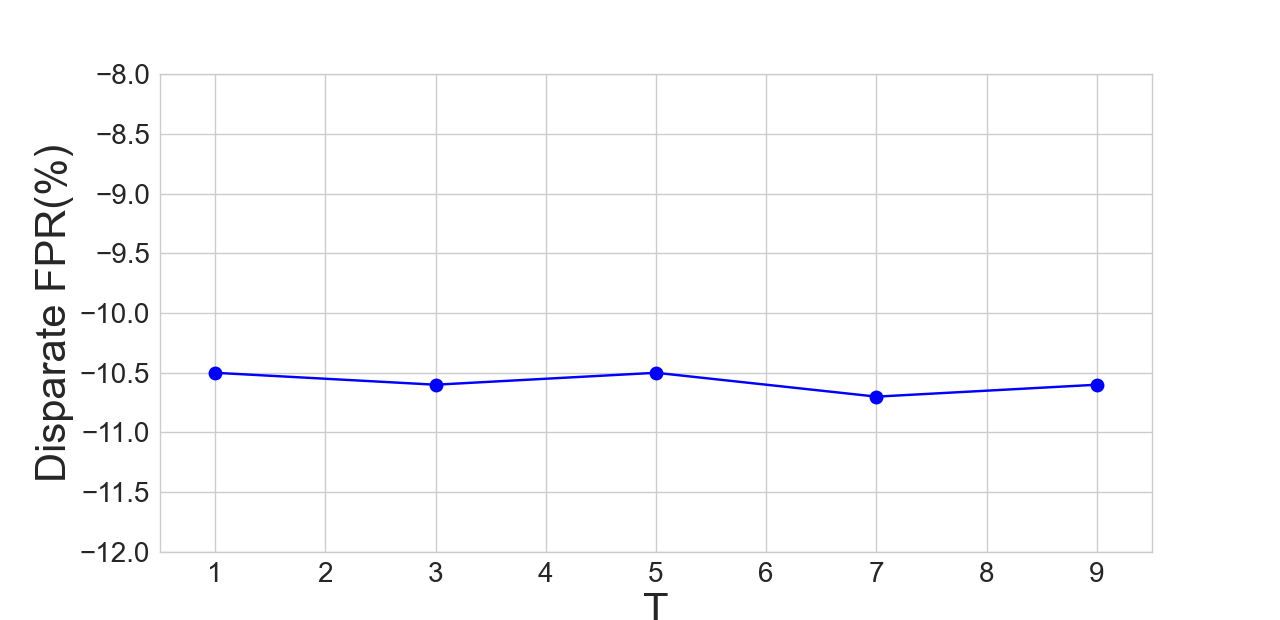}}
\subfloat[Disparate FNR vs. T]{\label{fig:d3}\includegraphics[width=40mm,height=1.2in]{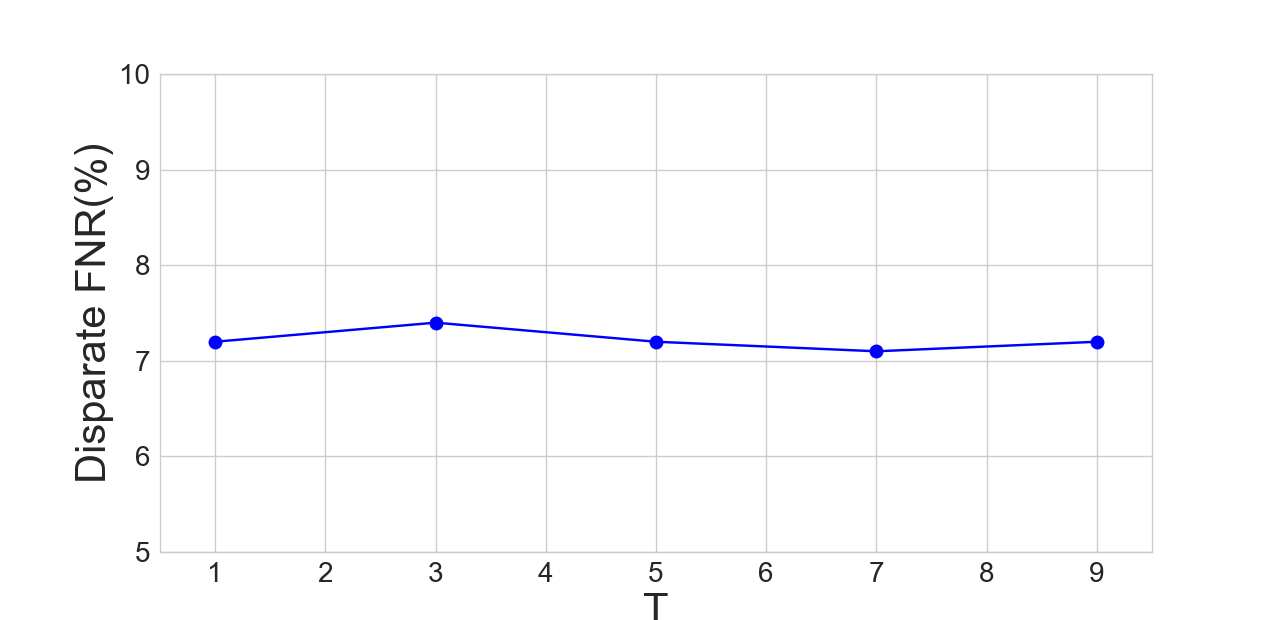}}
\caption{Sensitivity of metrics to the change of $T$ on Adult dataset}\label{fig:sens2}
\end{figure}

\subsubsection{Ablation test for assigning weights}\label{sec:ablation_} We try to assign weights only to the minority group. We could not close the Wasserstein distance gap and assign weights only to the majority group. Assigning weights to both groups could achieve similar results as our method. However, we might need an additional statistical test to claim so.

\subsubsection{Training on Adult Dataset and German Credit Dataset}\label{sec:training-tabular}
Since the tabular datasets have relatively lower dimensions than the image datasets, we could avoid using feature extractors. However, we use one-hot encoding to deal with the categorical features in the latent space. We could have a better continuity for the Wasserstein distance approximation. From Figure \ref{fig:sens2}, we can see that the metrics are not sensitive to the change of $T$ for the Adult dataset. Figure \ref{fig:ablation2} shows the sensitivity of different distance measures on the Adult dataset.

For the feature extractor $F_\phi$ and the classifier $C_\theta$, we also apply fully connected layers. For the discriminator $D$, we use the same architecture in \cite{gulrajani2017}, without the last sigmoid function. We apply SGD algorithm with a  momentum of 0.9 to update $\phi$ and $\theta$. The learning rate of $\theta$\ is ten times that of $\phi$. 
$\theta_D$ is updated by Adam algorithm with a learning rate 0.0001. Following \cite{gulrajani2017}, we adjust the
learning rate $\eta$ of $\theta$ by $\eta = \frac{0.01}{(1+10p)^{-0.75}}$, where $p$ is the training progress linearly changing from 0 to 1. We update $\phi$ and $\theta$
for 2 steps then update $\theta_D$ for 1 step.

\subsubsection{Multi-categorical sensitive attribute situation}

It is straight-forward to extend our method to handle a multi-categorical sensitive attribute or multi-sensitive attributes by using one subgroup as reference group and reweighing other subgroups alternatively (and in turn) to reach a state of demographic parity. We also use Adult dataset to demonstrate this. However, we choose $race$ here as the sensitive attribute. $race$ is $\{'Amer-Indian-Eskimo':0,
 'Asian-Pac-Islander':1,
 'Black':2,
 'Other':3,
 'White':4\}$ in the dataset. We use $'Asian-Pac-Islander'$ as the reference subgroup and reweighs samples from other subgroups. We report the disparate impact between the subgroup $'White'$ and $'Black'$. Before and after applying our method, the disparate impact is 15.1\% and 1.7\% and the accuracy is 83.1\% and 82.8\%.

\end{document}